\documentclass[10pt,twocolumn,letterpaper]{article}

\usepackage{3dv}
\usepackage{times}
\usepackage{epsfig}
\usepackage{graphicx}
\usepackage{amsmath}
\usepackage{amssymb}
\usepackage{multirow}
\usepackage{url}

\threedvfinalcopy 

\def\threedvPaperID{196} 

\ifthreedvfinal\fi

\begin{document}
	
	\title{ SphereDepth: Panorama Depth Estimation from Spherical Domain }


	\author{Qingsong Yan\textsuperscript{1,2}, 
	Qiang Wang\textsuperscript{3}, 
	Kaiyong Zhao\textsuperscript{4}, 
	Bo Li\textsuperscript{2}, \\
	Xiaoweo Chu\thanks{Corresponding author} \textsuperscript{,2,5}, 
	Fei Deng$^{*}$\textsuperscript{,1}\\
	\textsuperscript{1} Wuhan University, Wuhan, China \\
	\textsuperscript{2} The Hong Kong University of Science and Technology, Hong Kong SAR, China\\
	\textsuperscript{3} Harbin Institute of Technology (Shenzhen), Shenzhen, China\\
	\textsuperscript{4} XGRIDS, Shenzhen, China \\
	\textsuperscript{5} The Hong Kong University of Science and Technology (Guangzhou), Guangzhou, China\\
	{\tt\small  yanqs\_whu@whu.edu.cn, qiang.wang@hit.edu.cn, kyzhao@xgrids.com, bli@cse.ust.hk } \\
	{\tt\small xwchu@ust.hk, fdeng@sgg.whu.edu.cn }
    }
    

\maketitle
\thispagestyle{empty}

\begin{abstract}
	
	The panorama image can simultaneously demonstrate complete information of the surrounding environment and has many advantages in virtual tourism, games, robotics, etc. However, the progress of panorama depth estimation cannot completely solve the problems of distortion and discontinuity caused by the commonly used projection methods. 
	This paper proposes SphereDepth, a novel panorama depth estimation method that predicts the depth directly on the spherical mesh without projection preprocessing. 
	The core idea is to establish the relationship between the panorama image and the spherical mesh and then use a deep neural network to extract features on the spherical domain to predict depth. To address the efficiency challenges brought by the high-resolution panorama data, we introduce two hyper-parameters for the proposed spherical mesh processing framework to balance the inference speed and accuracy.
	Validated on three public panorama datasets, SphereDepth achieves comparable results with the state-of-the-art methods of panorama depth estimation. 
	Benefiting from the spherical domain setting, SphereDepth can generate a high-quality point cloud and significantly alleviate the issues of distortion and discontinuity. Our code are avilable at \url{https://github.com/Yannnnnnnnnnnn/SphereDepth}.
	
\end{abstract}

\begin{figure}[t]
	
	\begin{center}
		\includegraphics[width=1.0\linewidth]{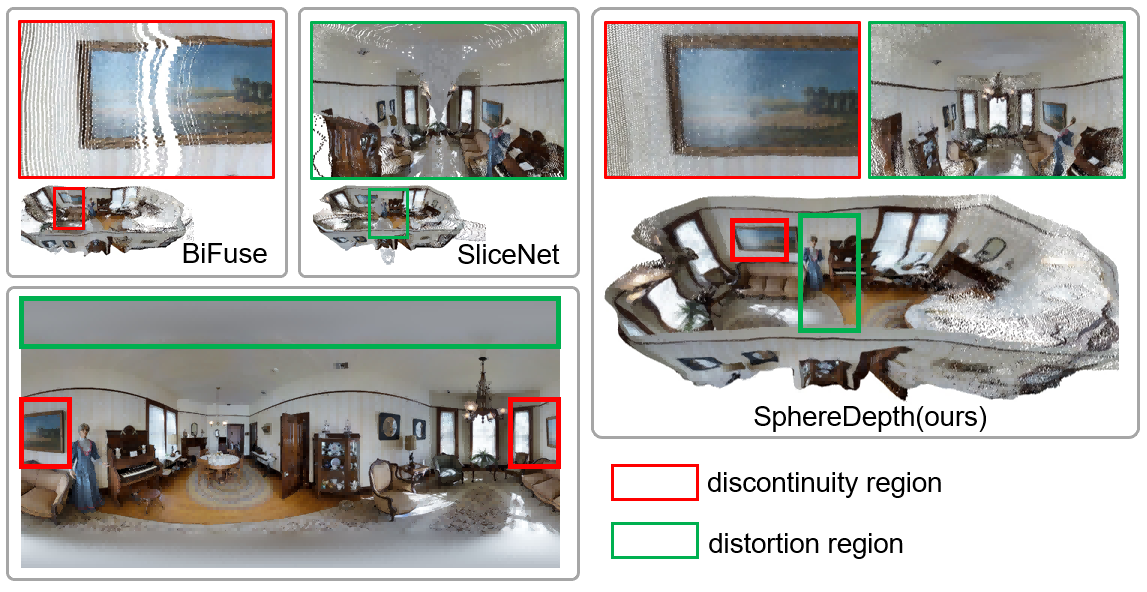}
	\end{center}
	
	\caption{
		\textbf{Point Cloud}
		SphereDepth can generate point clouds without distortion and discontinuity. BiFuse  \cite{wang2020bifuse} and SliceNet  \cite{pintore2021slicenet} suffer from discontinuity and distortion caused by equirectanglar projection.
	} 
	
	\label{fig:compare}
	
\end{figure}

\section{Introduction}

Depth estimation attempts to infer the distance between the objects and the camera in the captured scene, which is a classical but important vision problem for many upstream 3D applications, such as scene reconstruction, semantic understanding, robotics navigation, etc. Traditionally, researchers focus on the problem of monocular depth estimation \cite{monodepth2017}, which learns the traditional 2D content captured by typical pinhole projection model-based cameras. To obtain the novel information (360$^\circ$ sensing) of a 3D scene, they integrate those single views based on multi-view consistency \cite{gmvs2019,yao2018mvsnet} and structure-from-motion (SfM)  \cite{sfm2018,schonberger2016structure}. However, as the consumer-level 360$^\circ$ cameras are becoming more popular and rapidly developed in recent years, inferring the depth information from a panorama image naturally attracts the community. The panorama camera has a field-of-view (FoV) of 360$^\circ$ and can acquire a comprehensive view in just one shooting spot.

However, as the best way to display the panorama image is using the sphere (a non-Euclidean space), the existing studies of monocular depth estimation based on convolutional neural networks (CNNs) \cite{eigen2014depth,laina2016deeper,zhou2017unsupervised} cannot be directly applied to its depth task. Therefore, the most popular solution is using some projection methods to convert the panorama image into a standard 2D image, such as equirectangular projection and cube map projection. Equirectangular projection provides a wide FoV mimicking a peripheral vision but introduces distortion  \cite{sun2019horizonnet,su2017learning,zioulis2018omnidepth}, while cube map projection provides a smaller and non-distorted FoV mimicking the foveal vision but introduces discontinuity  \cite{cheng2018cube,jiang2021unifuse,jin2020geometric}. Overall speaking, those methods based on projection will inevitably suffer more or less from distortion and discontinuity and bring systematic errors to the upstream visual perception tasks. 

To this end, this paper proposes a novel panorama depth estimation method called SphereDepth, that directly predicts the depth map in the spherical domain without any projection. SphereDepth uses a spherical mesh to approximate the sphere and treats each triangle in a spherical mesh as a pixel in an image. After assigning each triangle with an RGB value, we can use a customized mesh convolution kernel to extract the features for depth estimation. We then construct a mesh-based convolution to directly perform depth estimation on a spherical mesh. Fig. \ref{fig:compare} compares the point cloud generated by SphereDepth with the existing state-of-the-art algorithm BiFuse \cite{wang2020bifuse} and SliceNet  \cite{pintore2021slicenet}. From the enlarged parts of the point cloud, we observe that SphereDepth provides accurate and smooth results while BiFuse and SliceNet exhibit discontinuity and distortion.

However, representing a high-resolution panorama image needs a spherical mesh with high resolution and more triangles. Unlike the perspective image organized in a 2D pixel matrix, the spherical mesh uses triangles to represent each pixel. The topological complexity of the spherical mesh exponentially grows with the number of triangles increasing, which degrades the computing efficiency and increases the memory footprint.

To address this challenge, we split the \textbf{spherical resolution} (SR) into two types of resolutions: the \textbf{mesh resolution} (MR), which indicates the number of triangles on a spherical mesh, and the \textbf{triangle resolution} (TR), which represents how many pixels each triangle has. On the one hand, a higher MR can represent a higher resolution panorama image with more details and naturally lead to larger resource consumption. On the other hand, compared to increasing MR, a high TR can enhance the local features of a mesh triangle with much less computational demand and memory demand. In practice, we can tune MR and TR in the spherical domain to balance the accuracy and processing speed of the deep network. Our contributions are summarized as follows.

\begin{itemize}
	\item SphereDepth is an end-to-end network to infer the panorama depth map in the spherical domain, which fundamentally solves the problem of distortion and discontinuity caused by projection methods;
	\item We sample the features from a panorama image in the spherical domain and propose customized operations to tackle spherical features for panorama depth estimation. We also propose two hyper-parameters for SphereDepth, MR and TR, to balance the efficiency and accuracy;
	\item Superior to those existing methods, SphereDepth achieves comparable accuracy in panorama depth estimation and generates a much higher quality point cloud. SphereDepth suggests a new direction to apply the sphere mesh convolution to the panorama image instead of traditional convolution networks.
\end{itemize}

\begin{figure*}
	
	\begin{center}
		\includegraphics[width=1.0\linewidth]{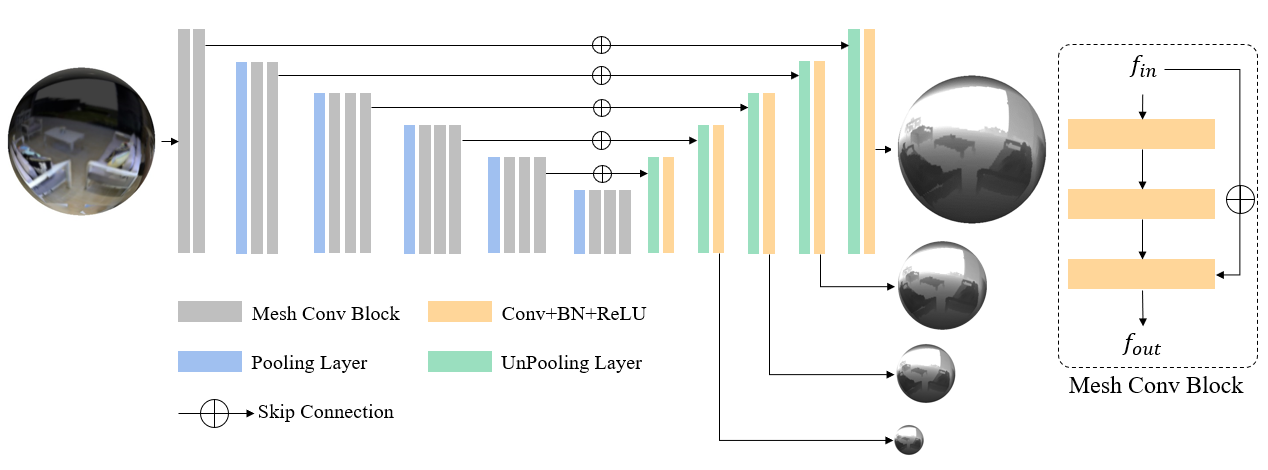}
	\end{center}
	
	\caption{
		\textbf{SphereDepth}
		uses mesh convolution to estimate depth on triangles of the spherical mesh to avoid distortion and discontinuity, instead of directly using traditional 2D convolution, such as BiFuse \cite{wang2020bifuse} and SliceNet \cite{pintore2021slicenet}.
	}
	
	\label{fig:pipeline}
	
\end{figure*}

\section{Related}


\paragraph{Perspective Depth Estimation}

Perspective Depth Estimation is a hot topic in computer vision. 
Eigen \cite{eigen2014depth} proposed the first deep learning network that can estimate a depth map directly from an RGB image. Laina \cite{laina2016deeper} uses ResNet \cite{he2016deep} as the encoder and skip connections to merge high-level and low-level features. DORN \cite{fu2018deep} divides the space into discrete intervals and treats the depth estimation problem as a classification problem. DenseDepth \cite{alhashim2018high} introduces more loss functions to obtain a smoother depth map. Recently, Midas \cite{ranftl2020towards} and BoostedDepth \cite{miangoleh2021boosting} have explored some strategies that can improve the quality of depth estimation, such as a mixture of different datasets and changing the image resolution. Meanwhile, Adabins  \cite{bhat2021adabins} uses ViT\cite{dosovitskiy2020image} to guess the best discrete depth distribution, and DPT \cite{ranftl2021vision} directly proposes a transformer network that can perform dense prediction.

All the above methods require a label for each pixel, which is difficult to satisfy in natural scenes. So some researchers turn to self-supervised methods. SfMLearner \cite{zhou2017unsupervised} uses the photo-metrics as a guide to train the network and predict the depth and the relative pose at the same time. MonoDepth2 \cite{godard2019digging} uses a robust error detection method to filter out dynamic areas and invisible areas, which is against the static scene assumption in the self-supervised method. SfMLearner-SC \cite{bian2021unsupervised} tries to solve the problem of scale drift in the self-supervised method by encouraging the consistency of depth maps. TrainFlow \cite{zhao2020towards} realizes the unreliable of the pose net and uses an optical flow to calculate the relative pose to improve the robustness. CVD \cite{luo2020consistent} uses colmap \cite{schonberger2016structure} to obtain the pose directly and apply a deep network to predict the depth by the photo consistency.

However, all methods mentioned above only work for perspective cameras and ignore distortion. Thus, they cannot be directly adapted to the 360$^\circ$ panorama images.

\begin{table}
	\begin{center}
		\caption{The Resolution Look-Up Table}
		\label{tab:look_up_resolution}
		\begin{tabular}{cc}
			\hline\noalign{\smallskip}
			Image Resolution (IR) & Spherical Resolution (SR) \\
			\noalign{\smallskip}
			\hline
			\noalign{\smallskip}
			32$\times$64    & 4 \\
			64$\times$128   & 5 \\
			128$\times$256  & 6 \\
			256$\times$512  & 7 \\
			512$\times$1024 & 8 \\
			\hline
		\end{tabular}
	\end{center}
\end{table}

\paragraph{Panorama Depth Estimation}

Although the progress of perspective panorama depth estimation is enormous, the study of panorama depth estimation is still at a certain early age. Recent studies  \cite{zioulis2018omnidepth,wang2020bifuse,pintore2021slicenet,sun2021hohonet} still rely on traditional deep neural networks (DNNs) in the task of panorama depth estimation. As DNNs can only work on the 2d image, equirectangular projection and cubemap projection are applied to the panorama image before processing.

Equirectangular projection allows all surrounding information to be observed from a single 2D image while introducing distortion and discontinuity at the boundary. Early studies, such as LayoutNet \cite{zou2018layoutnet} and HorizonNet \cite{sun2019horizonnet}, directly applied DNNs to the equirectangular projection format and ignored distortion and discontinuity. 
Some recent studies applied special modules in DNNs, such as deformable convolution in SphereNet \cite{coors2018spherenet} and Corners-for-Layout \cite{fernandez2020corners}, and the customized layer in OmniDepth \cite{zioulis2018omnidepth}. 
SliceNet \cite{pintore2021slicenet} followed the idea of OmniDepth with a specially designed network structure that extends the reception field of the convolution kernel. 
Besides, Su \cite{su2017learning} proposed to train an independent sub-network to correct the distortion in the feature maps generated by the backbone structure.

Unlike equirectangular projection, cubemap projection transforms the spherical content into six different perspective faces of a cube. One then can adopt the traditional DNN structures for monocular depth estimation to each face and re-project them back to the spherical domain. However, the cubemap projection introduces discontinuity at the boundary of the cube. Wang \cite{wang2018self} directly used cubemap to estimate the depth map ignoring the discontinuity. Cheng \cite{cheng2018cube} and BiFuse \cite{wang2020bifuse} proposed special padding methods to connect those perspective faces.
In particular, BiFuse \cite{wang2020bifuse} integrated features from  equirectangular projection to mitigate discontinuity. Jin \cite{jin2020geometric} combines the depth and the geometric structure to improve the quality of the depth map. 

To eliminate the disadvantages of the above two projection methods, some recent studies attempted to process the panorama image in the spherical domain and apply a designed convolution in some 2D vision tasks, such as semantic segmentation in  \cite{jiang2018spherical,lee2019spherephd,zhang2019orientation,eder2020tangent}, object detection/classification in \cite{lee2019spherephd,cohen2018spherical} and layout estimation \cite{pintore2021deep3dlayout}.
However, it still lacks discussions on applying spherical convolution to inferring 3D information, such as depth, from a panorama image, because of the heavy computation burden.
We propose a new panorama depth estimation using the spherical mesh and follow the setting of the SubdivNet \cite{hu2021subdivision}, which is more efficient than MeshCNN \cite{hanocka2019meshcnn} and MeshNet \cite{feng2019meshnet}. Our experiments show that our method can achieve a comparable result with state-of-the-art results.

\section{Method}

We present the details of the proposed method in this section. We first discuss how to represent the panorama image by the spherical mesh in Section \ref{sec:spherical_mesh} and then show how to build an end-to-end network that can predict the depth map in the spherical domain in Section \ref{sec:mesh_conv_pool} and Section \ref{sec:network_and_loss}.

\subsection{ Spherical Mesh }
\label{sec:spherical_mesh}

Instead of applying the equirectangular projection or cubemap projection to the panorama image, we use the spherical mesh to process the panorama image in the spherical domain. The spherical domain can help us avoid discontinuity and distortion caused by projections. A popular method to represent the panorama image in the spherical domain is based on the icosahedron spherical mesh (ISM) \cite{eder2020tangent}, which can generate a higher resolution of spherical mesh to approximate a sphere by loop-subdivision \cite{hu2021subdivision}.		

As most panorama images are stored in the equirectangular projection, we have to represent them by the spherical mesh. Considering the fundamental element of the spherical mesh is the triangle, a straightforward method is to project the triangle's center $p=(x,y,z)$ to the image plane and sample color on the image plane based on the projected point $i=(u,v)$. We can derive the geometric relationship between the 3D point $p$ and 2D point $i$ on an image with the image resolution (IR) $(W, H)$ by Eq. \ref{eq:uv_xyz}.

\begin{align}
    \label{eq:uv_xyz}
	\begin{cases}
		u = ( 1 + atan2(y,x)/\pi ) \times W/2 \\
		v = ( 0.5 + atan2(z,\sqrt{x^2+y^2})/\pi ) \times H
	\end{cases}
\end{align}

When a triangle only displays one pixel, it is easy to realize that a higher IR needs a higher spherical resolution (SR) that determines the number of triangles by $20 \times 4^{SR}$. Following Tangent \cite{eder2020tangent}, we can list out the IR-SR relationship in Table \ref{tab:look_up_resolution}. For a given IR, we must use the corresponding SR to get the best results. However, we could not process spherical mesh with lots of triangles efficiently \cite{hu2021subdivision}. Therefore, instead of only displaying one pixel by a triangle,  we split the SR into mesh resolution (MR) and triangle resolution (TR).

\begin{figure}
	
	\begin{center}
		\includegraphics[width=.8\columnwidth]{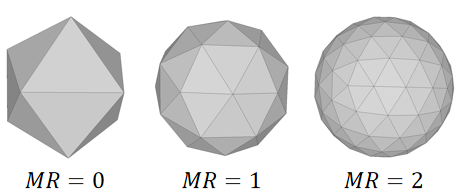}
		\includegraphics[width=.8\columnwidth]{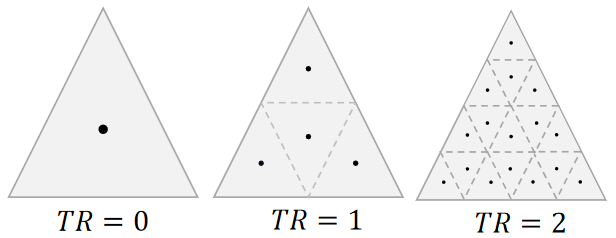}
	\end{center}
	
	\caption{
		\textbf{Spherical Resolution} includes $MR$ (Mesh Resolution) and $TR$ (Triangle Resolution).
	}
	
	\label{fig:resolution}
	
\end{figure}

\begin{figure}
	
	\begin{center}
		\includegraphics[width=.8\columnwidth]{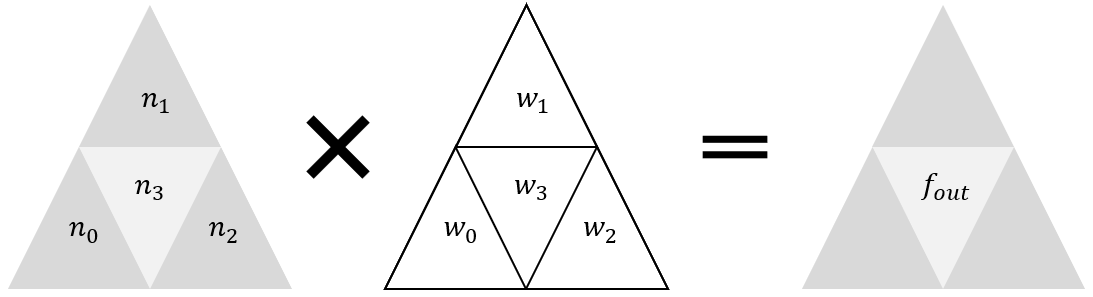}
		\includegraphics[width=.8\columnwidth]{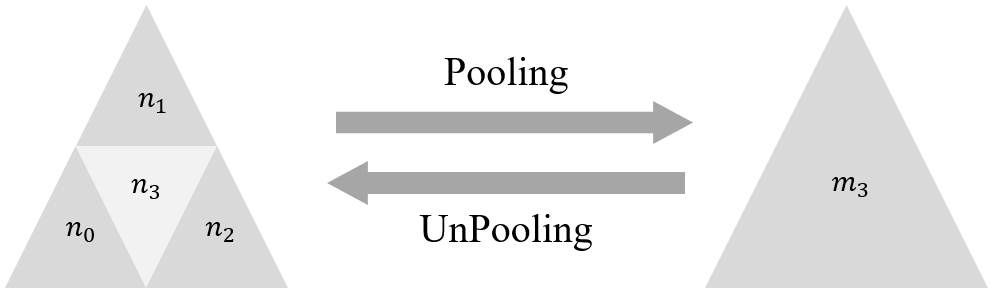}
	\end{center}
	
	\caption{
		\textbf{Mesh Convolution and Pooling/UnPooling} The first row shows the mesh convolution operation and the second row shows the pooling/unpooling operation. 
	}
	
	\label{fig:mesh_kernel_pooling}
	
\end{figure}

\paragraph{Mesh Resolution}
Mesh resolution (MR) refers to the number of times the ISM is loop-subdivided and decides the number of triangles on a spherical mesh. The higher of MR, the closer is between the spherical mesh and the standard sphere. Tangent \cite{eder2020tangent} uses the Surface Area Ratio to measure the difference and finds out that it is almost equal to 1 when $MR>3$. However, the computational complexity of the mesh convolution is linear to MR, and higher MR will bring a more complex topological relationship and make it difficult to process. Therefore, we have to find out a proper MR.

\paragraph{Triangle Resolution} 
Triangle resolution (TR) means the number of points sampled within a triangle. TR is also achieved by loop-subdivision, but we only keep triangles' center and ignore topological relationships. The relation between the number of points and TR is $4^{TR}$. During the training and inference procedure, we directly concatenate all the features coming from TR together in a fixed order and feed them to the network simultaneously. Fig. \ref{fig:resolution} shows three triangles with different triangle resolutions, which shows how to get a spherical mesh with higher resolution and how to generate more points in a triangle.

\subsection{ Mesh Convolution and Pooling/UnPooling }
\label{sec:mesh_conv_pool}

After representing a panorama image in the spherical domain by a spherical mesh, we design a customized mesh convolution kernel to extract the spherical features and follow pooling/unpooling from SubdivNet \cite{hu2021subdivision}. 

Unlike the 2d pixels, which can directly access neighboring pixels by coordinates, the triangles in the spherical mesh cannot find their adjacent triangles by coordinates. The only way to gather this kind of information is through the topological relationship of the mesh, called face adjacent face (FAF). Considering that the number of FAF of a triangle is not fixed, we simplify this by defining adjacent triangles as the subset of the FAF with edges in common to reduce the number of adjacent triangles to 3, as Fig. \ref{fig:mesh_kernel_pooling} shows that there are only three triangles around $n_3$. We define a simple mesh convolution kernel in Eq. \ref{eq:conv} with a fixed number of adjacent triangles, in which $f_{out}$ is the output features, $n_{i} (i=0,1,2,3)$ are the input features, and $w_{i} (i=0,1,2,3)$ are kernel parameters needed to be learned (Bias is used, but not visualized here).

\begin{align}
    \label{eq:conv}
	f_{out}=w_0n_{0}+w_1n_{1}+w_2n_{2}+w_{3}n_{3}
\end{align}

The mesh pooling/unpooling kernel is also an indispensable module for building a network of encoder-decoder paradigms. Due to the unique nature of the spherical mesh, we can directly merge subdivided triangles into one triangle, and the reverse operation is unpooling. As Fig. \ref{fig:mesh_kernel_pooling} shows, four triangles $n_0,n_1,n_2,n_3$ merge into one triangle $m_3$ by pooling, and one triangle $m_3$ can be split into four triangle $n_0,n_1,n_2,n_3$ by unpooling. Furthermore, the pooling will change the spherical resolution of the feature map in the spherical domain, which means that the number of pooling layers cannot exceed MR. Throughout our paper, we use max-pooling.

\subsection{ Network and Loss Function }
\label{sec:network_and_loss}

\paragraph{Network}

We construct our SphereDepth based on UNet Structure \cite{ronneberger2015u}, as shown in Fig. \ref{fig:pipeline}. 
The encoder part consists of five pooling layers to increase the reception field and 2 or 3 mesh convolution blocks in each layer to extract robust features. In each mesh convolution block, we follow the design of ResNet  \cite{he2016deep} and stack three Conv Layers together with a residual connection. On the decoder side, we use five unpooling layers to increase the resolution of the feature map and concatenate them with feature maps from the encoder by skip-connection. Benefiting from the design of UNet \cite{ronneberger2015u}, SphereDepth can generate depth maps with different spherical resolutions. The input and output of the SphereDepth are represented in the spherical domain. For an input with $MR=m$ and $TR=t$, the shape of input is $(1,20 \times 4^m,c \times 4^t)$ and the shape of output is $(1,20 \times 4^{m-s},4^t)$. $c$ is the channel number of the image, and $s$ is the output stage.

\paragraph{Log Loss Function} 

We utilize the multi-resolution depth maps predicted by SphereDepth and construct a multi-scale loss as Eq. \ref{eq:loss} shows, where $r$ is the spherical resolution of the depth map, $l_r$ is the weight set for each resolution, $R$ is the max spherical resolution, $V$ is the set of valid pixels, $N_r$ is the number of valid pixels in the $r$ resolution level, $gt$ is the ground truth, and $pr$ is the predicted depth map. During training, we define valid pixels by gt depth maps whose values are inside a preset depth range.

\begin{align}
    \label{eq:loss}
	loss= \sum_{r}^{R}{\sum_{p\in V}l_r{\frac{|log(gt_r(p)) - log(pr_r(p))|}{N_r} }}.
\end{align}

\section{Experiments}

\subsection{Datasets }

We test our method on three popular panorama datasets, Stanford2D3D \cite{armeni2017joint}, Matterport3D \cite{chang2017matterport3d} and 360D \cite{zioulis2018omnidepth}. Details of each dataset are introduced below.

\paragraph{Stanford2D3D} Stanford2D3D \cite{armeni2017joint} is collected from the real world. It contains 1413 panoramas from three types of buildings, including six large-scale indoor areas. We resize the panorama images and depth maps into 512$\times$1024 and follow the official splits for training and testing.

\paragraph{Matterport3D} Matterport3D \cite{chang2017matterport3d} is a real-world dataset, which has 10800 panorama images from 90 different rooms captured by Matterport’s Pro 3D Camera. We use the official split, including 61 rooms for training and 29 rooms for testing. We resize the panorama image to 512$\times$1024. 

\paragraph{360D} 360D \cite{zioulis2018omnidepth} is a large synthetic dataset rendered by OmniDepth \cite{zioulis2018omnidepth} using Ray Tracing Method. 360D uses texture models from four different datasets (including Stanford2D3D and Matterport3D) and generates 35977 panorama images. From Omnidepth, 34679 of this dataset are used for training, and the rests are for testing. Unlike the real-world datasets, the resolution of 360D is 256$\times$512.

\subsection{Implementation}

We implement SphereDepth by the Jittor \cite{hu2020jittor} framework, which is a high-performance deep learning framework based on JIT (just in time) compiling and meta-operators.
We set the batch size to 4 and use Adam \cite{kingma2014adam} optimizer with a learning rate of 4e-4. Considering the small size of Standford2D3D, we use the trained model from Matterport3D to fine-tune this dataset. 
We compare our method with FCRN \cite{laina2016deeper}, OmniDepth \cite{zioulis2018omnidepth}, BiFuse \cite{wang2020bifuse} and SliceNet \cite{pintore2021slicenet}. Following BiFuse and SliceNet, we use MAE, MRE, RMSE, RMSE(log) and $\delta$ as evaluation metrics and set the max depth value to $10$ meters for 360D and $16$ meters for Standford2D3D and Matterport3D. Details of those metrics are explained in the supplementary materials.

\subsection{Ablation Study}

In this section, we present the ablation study on the architecture settings of SphereDepth, including the loss function, convolution kernel, and spherical resolution. We use the largest 360D for all the ablation experiments to get robust results. Our default setting is using UNet \cite{ronneberger2015u} as an encoder with a log loss function, and the SR is $\{MR=5, TR=2\}$. We only modify part of the setting in each ablation experiment and keep the rest fixed.

\paragraph{Loss function}

To improve the performance of SphereDepth, we conduct experiments on three different loss functions, Log-loss, Absolute-loss, and Huber-loss \cite{wang2020bifuse}. Log-loss is calculated in the logarithmic domain and can pay more attention to the information in the closer area. Absolute loss directly calculates the difference, also known as L1 loss. Huber-loss combines L1 loss and L2 loss, which is more robust to outliers.
Table \ref{tab:choose_of_encoder_loss} shows the results of different settings. The Log-loss function gets the best results, and the worst is the Huber-loss function. Based on these results, in the later experiments, we choose UNet as the encoder of SphereDepth and the Log-loss function.

\begin{table}
	\begin{center}
		\caption{Ablation study of different loss functions on 360D.}
		\label{tab:choose_of_encoder_loss}
		\addtolength{\tabcolsep}{-0.2pt}
		\begin{tabular}{lccc}
			\hline\noalign{\smallskip}
			Loss & MRE$\downarrow$ & MAE$\downarrow$ & RMSE$\downarrow$  \\
			\noalign{\smallskip}
			\hline
			\noalign{\smallskip}
			
			log & \textbf{0.0700} & \textbf{0.1444} & \textbf{0.2765} \\
			abs & 0.0789 & 0.1540 & 0.2809 \\
			huber & 0.0771 & 0.1527 & 0.2814 \\ 
			
			\hline
		\end{tabular}
	\end{center}
\end{table}

\begin{table}
	\begin{center}
		\caption{Ablation study of the convolution kernel on 360D.}
		\label{tab:kernel_type}
		\begin{tabular}{lccc}
			\hline\noalign{\smallskip}
			Kernel & MRE$\downarrow$ & MAE$\downarrow$ & RMSE$\downarrow$ \\
			\noalign{\smallskip}
			\hline
			\noalign{\smallskip}
			Ours  & \textbf{0.0700} & \textbf{0.1444} & \textbf{0.2765}  \\
			SubdivNet\cite{hu2021subdivision} & 0.0936 & 0.1810 & 0.3131  \\
			
			\hline
		\end{tabular}
	\end{center}
\end{table}

\begin{table}
	\begin{center}
		\caption{Ablation study of spherical resolution on 360D.}
		\label{tab:choose_of_resolution}
		\addtolength{\tabcolsep}{-1.8pt}
		\begin{tabular}{cccccc}
			\hline\noalign{\smallskip}
			MR,TR & Time$\downarrow$ & GPU$\downarrow$ & MRE$\downarrow$ & MAE$\downarrow$ & RMSE$\downarrow$ \\
			\noalign{\smallskip}
			\hline
			\noalign{\smallskip}
			5,1 & \textbf{$\sim$60ms} & \textbf{$\sim$3.8G} & 0.0700 & 0.1464 & 0.2863 \\ 
			5,2 & $\sim$80ms & $\sim$4.2G & 0.0700 & 0.1444 & 0.2765 \\
			6,1 & $\sim$240ms & $\sim$14G & \textbf{0.0642} & \textbf{0.1311} & \textbf{0.2495} \\
			
			\hline
		\end{tabular}
	\end{center}
\end{table}

\paragraph{The Convolution Kernel}

We conduct another ablation experiment on the convolution kernel. The convolution kernel of SubdivNet \cite{hu2021subdivision} focuses on the geometry information of the mesh and uses a rotation-invariant kernel to process the features. In contrast, the convolution kernel of SphereDepth follows the principle of 2D convolution kernel and applies weighted aggregation to the input features. Notice that the original paper of SubdivNet only discusses the case of image classification. We modify the code of SubdivNet to fit our depth estimation pipeline accordingly. 
Table \ref{tab:kernel_type} shows the kernel of SphereDepth outperforms the kernel of SubdivNet with a lower computational.

\begin{figure}
	
	\centering
	
	\hspace{0.02\linewidth} RGB \hspace{0.14\linewidth} GT \hspace{0.13\linewidth} Ours \hspace{0.12\linewidth} PointCloud
	
	\includegraphics[width=.24\linewidth]{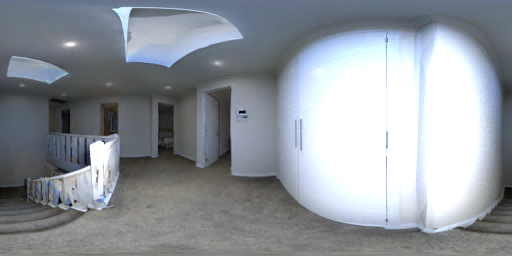}
	\includegraphics[width=.24\linewidth]{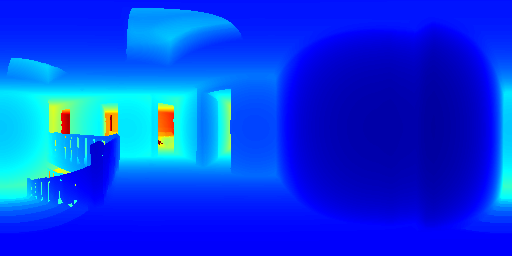}
	\includegraphics[width=.24\linewidth]{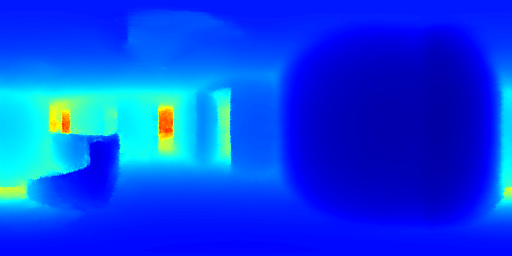}
	\includegraphics[width=.24\linewidth]{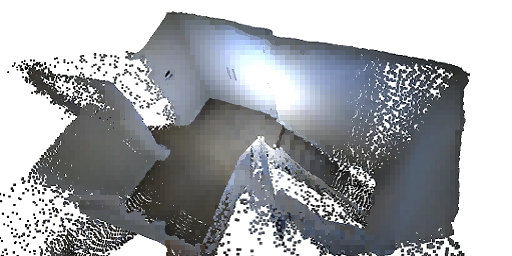}
	
	\includegraphics[width=.24\linewidth]{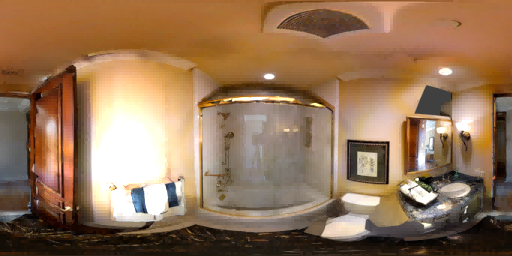}
	\includegraphics[width=.24\linewidth]{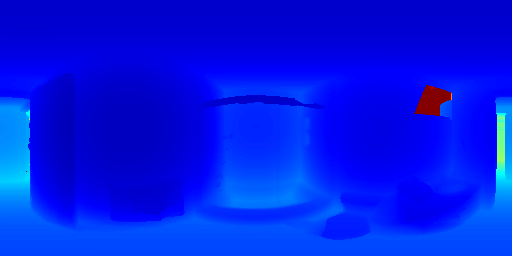}
	\includegraphics[width=.24\linewidth]{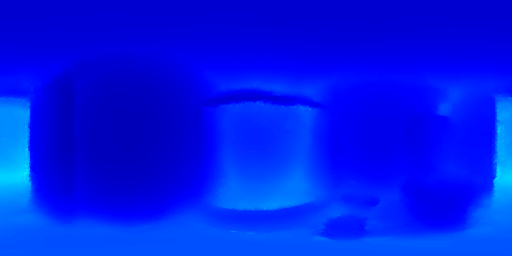}
	\includegraphics[width=.24\linewidth]{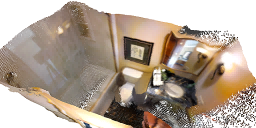}
	
	\caption{
		\textbf{Results of 360D}
		Invalid parts of the depth map are set to red. For better visualization of the point cloud, we remove the ceiling part. 
	}
	\label{fig:vis_3d60}
	
\end{figure}

\begin{figure}
	\centering
	
	\hspace{0.01\linewidth} RGB \hspace{0.15\linewidth}  GT \hspace{0.10\linewidth} SliceNet  \cite{pintore2021slicenet} 
	\hspace{0.07\linewidth} Ours 
	
	\includegraphics[width=.24\linewidth]{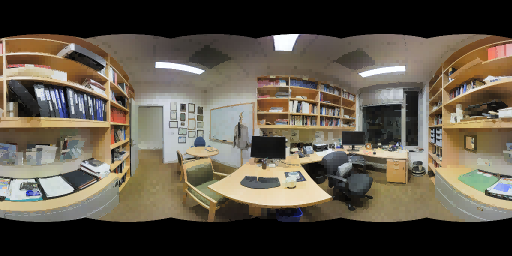}
	\includegraphics[width=.24\linewidth]{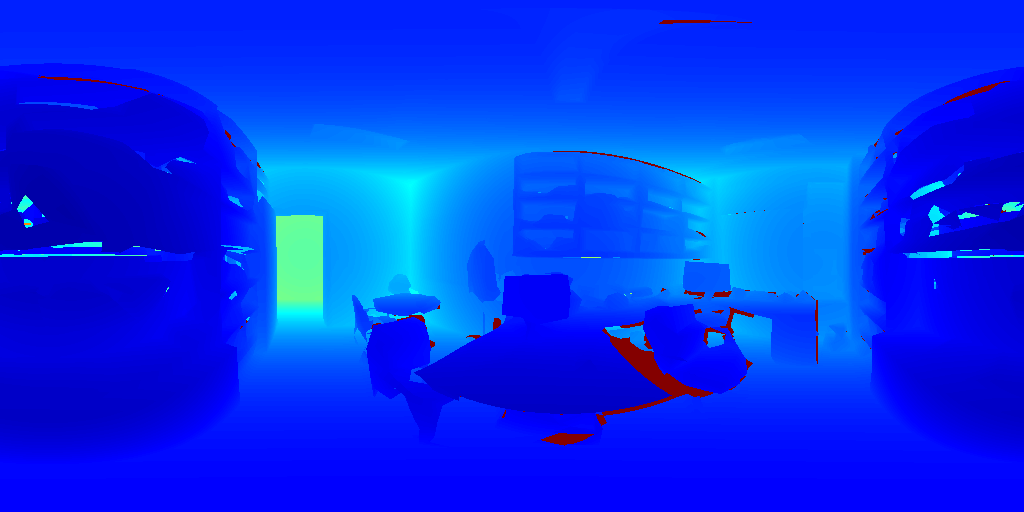}
	\includegraphics[width=.24\linewidth]{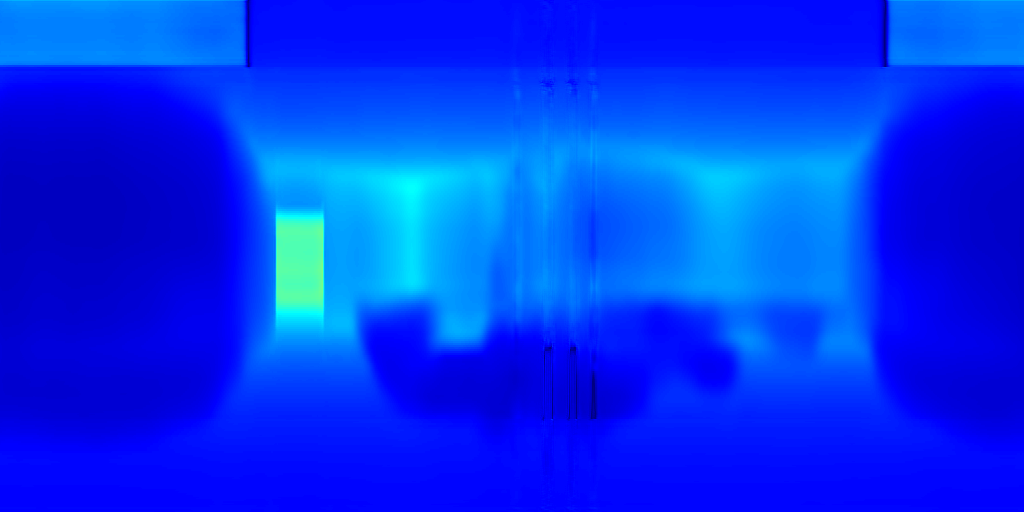}
	\includegraphics[width=.24\linewidth]{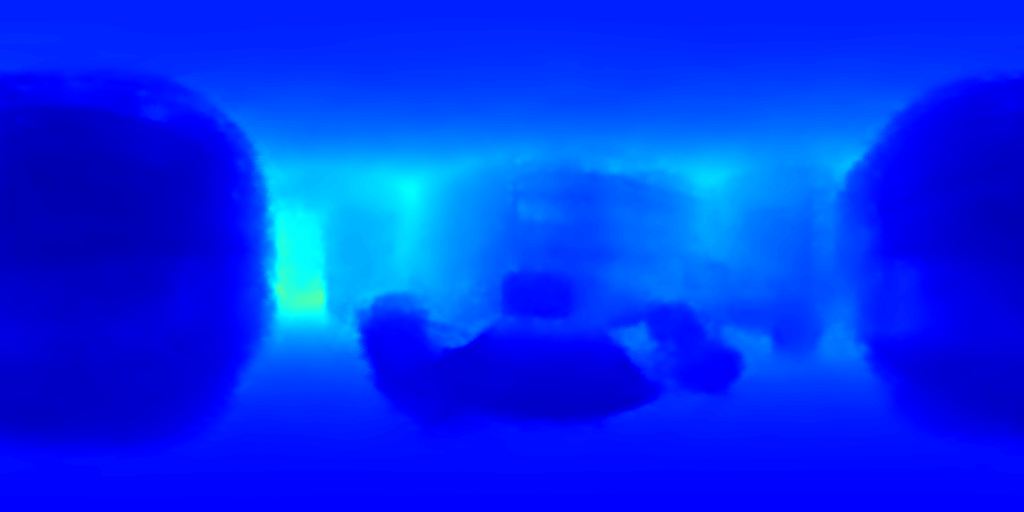}
	
	\includegraphics[width=.24\linewidth]{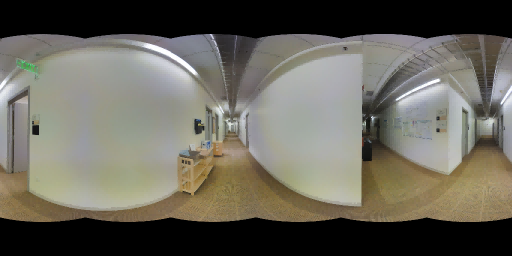}
	\includegraphics[width=.24\linewidth]{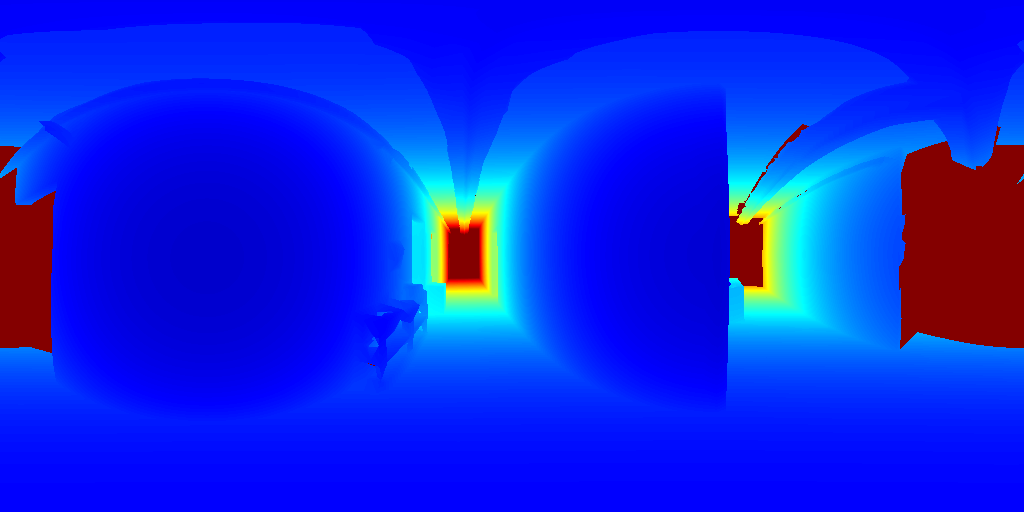}
	\includegraphics[width=.24\linewidth]{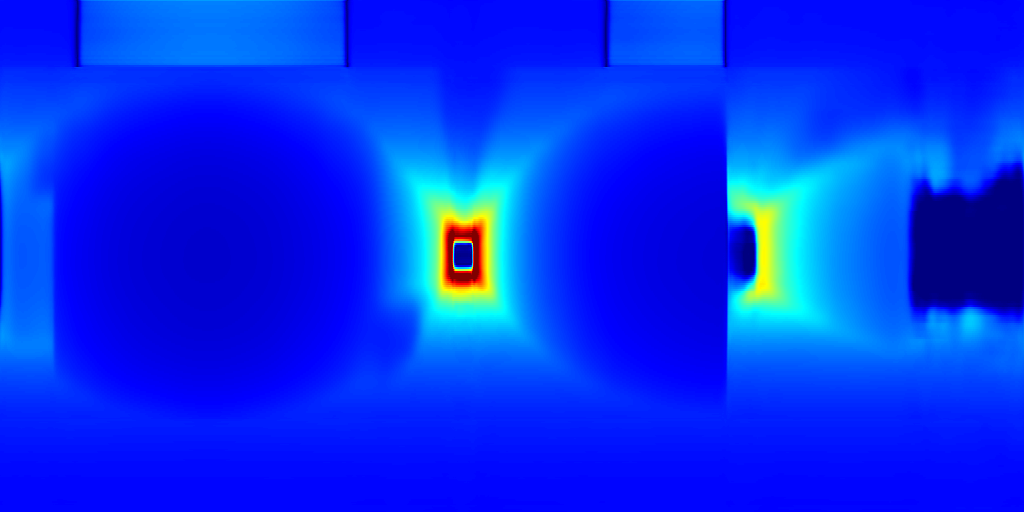}
	\includegraphics[width=.24\linewidth]{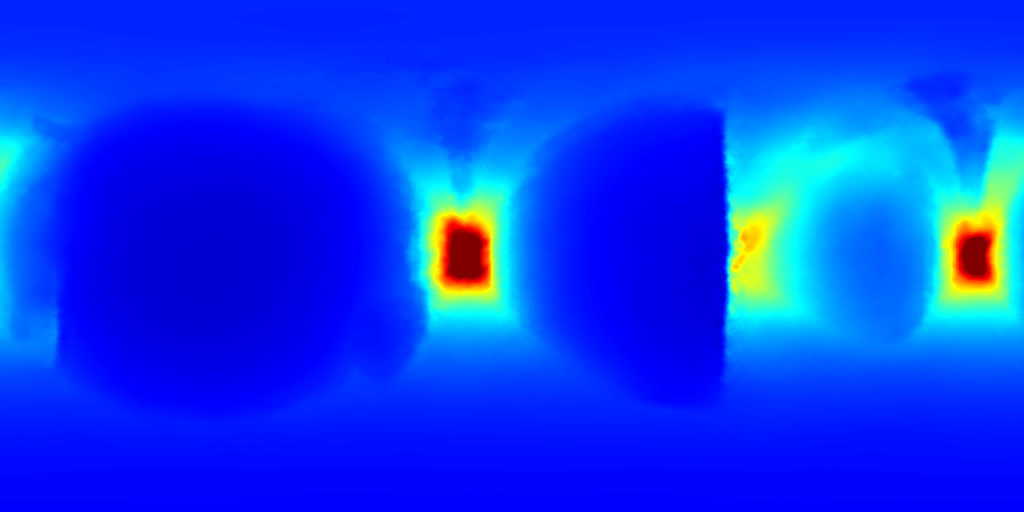}
	
	\caption{
		\textbf{Results of Standford2D3D}
		Invalid parts of the depth map are set to red. SphereDepth (ours) can generate reliable depth values in the pole regions, but SliceNet cannot predict the correct depth.
	}
	\label{fig:vis_2d3d}
	
\end{figure}

\begin{figure}
	\centering
	
	\hspace{0.01\linewidth}SliceNet \cite{pintore2021slicenet} 
	\hspace{0.30\linewidth}Ours 
	
	\includegraphics[width=.45\linewidth]{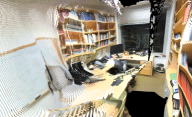}
	\includegraphics[width=.45\linewidth]{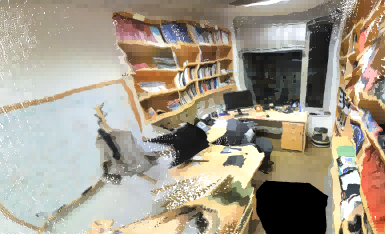}
	
	\vspace{0.01\linewidth}
	
	\includegraphics[width=.45\linewidth]{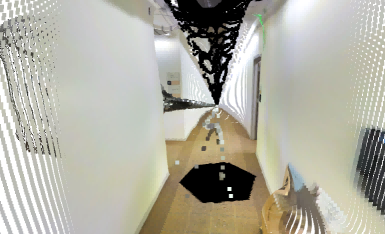}
	\includegraphics[width=.45\linewidth]{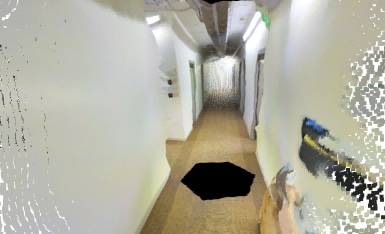}

	\caption{
		\textbf{PointCloud Results on Standford2D3D} SliceNet's depth map produces lots of noise in the point cloud in the pole regions.
	}
	
	\label{fig:vis_2d3d_pointcloud}
\end{figure}

\begin{figure*}
	\centering
	
	\hspace{0.02\linewidth} RGB \hspace{0.12\linewidth}  GT \hspace{0.12\linewidth} BiFuse \cite{wang2020bifuse} \hspace{0.10\linewidth} SliceNet \cite{pintore2021slicenet} \hspace{0.11\linewidth} Ours
	
	\includegraphics[width=.19\linewidth]{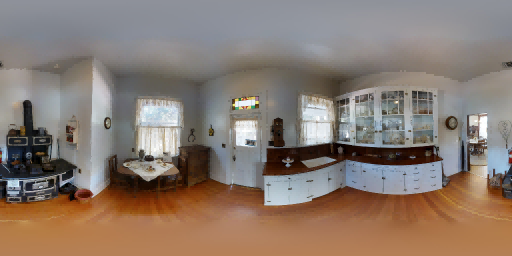}
	\includegraphics[width=.19\linewidth]{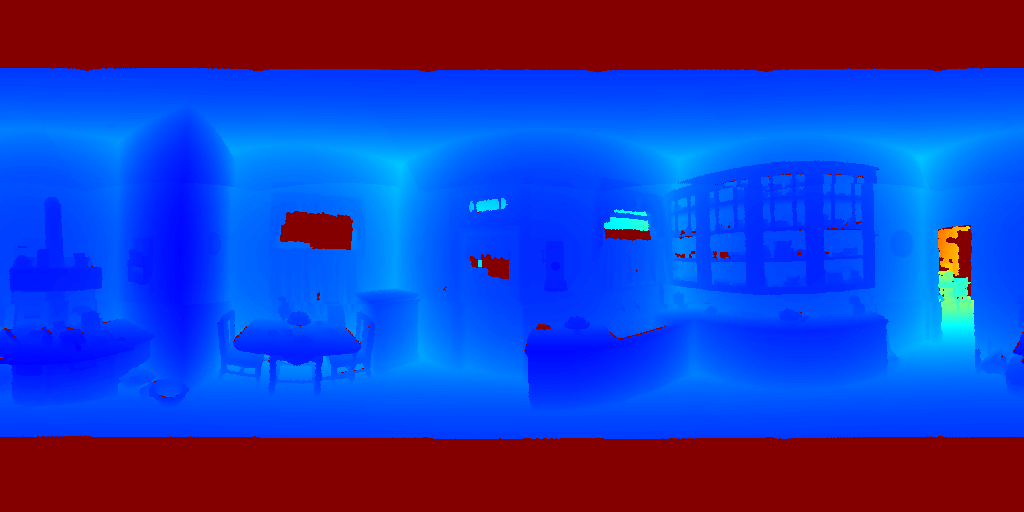}
	\includegraphics[width=.19\linewidth]{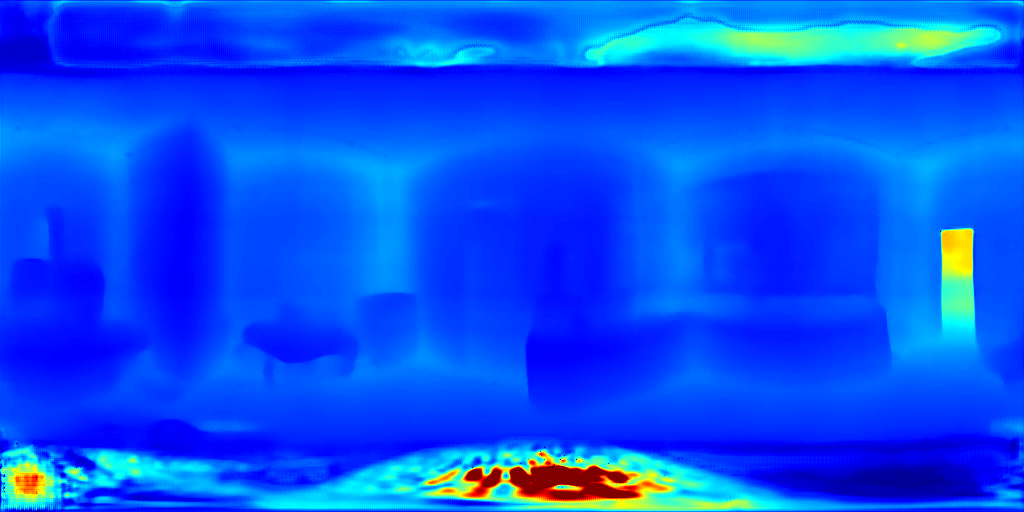}
	\includegraphics[width=.19\linewidth]{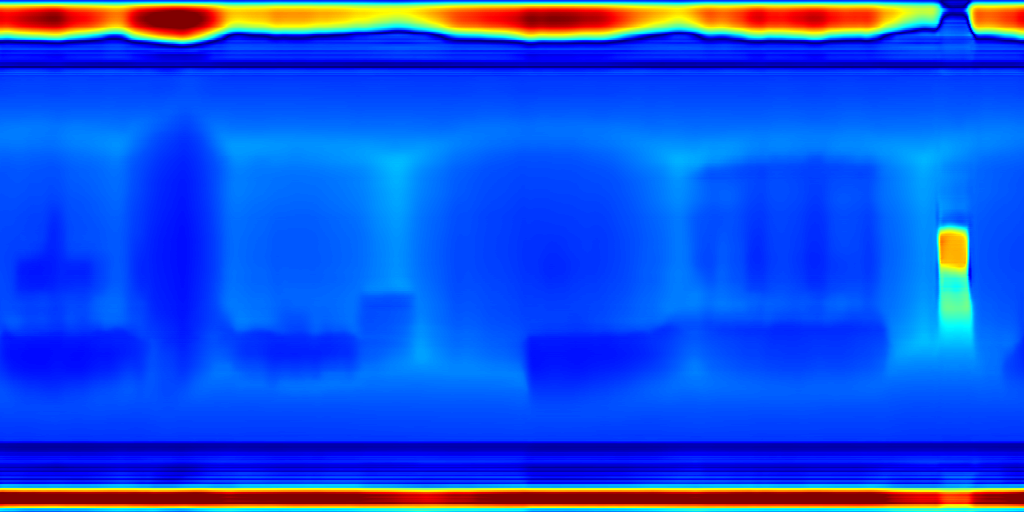}
	\includegraphics[width=.19\linewidth]{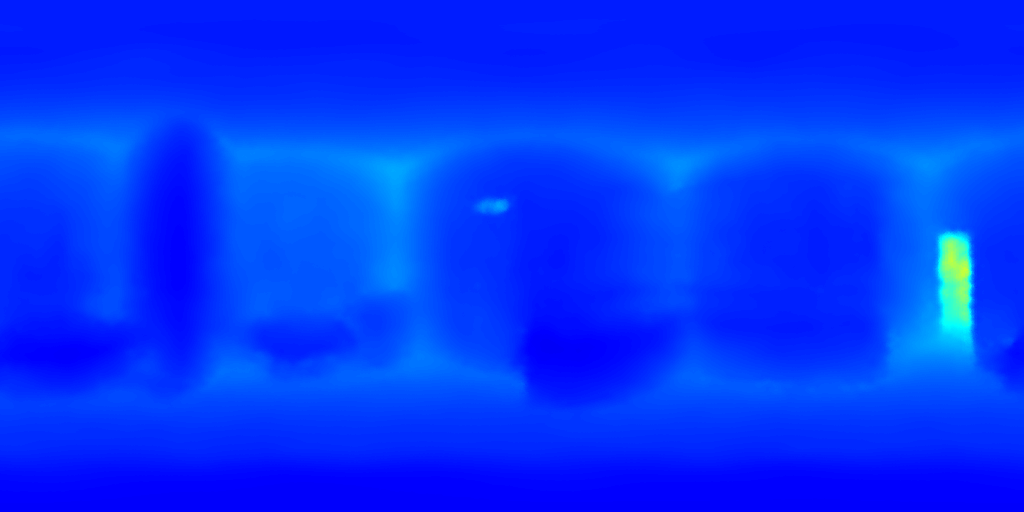}
	
	\includegraphics[width=.19\linewidth]{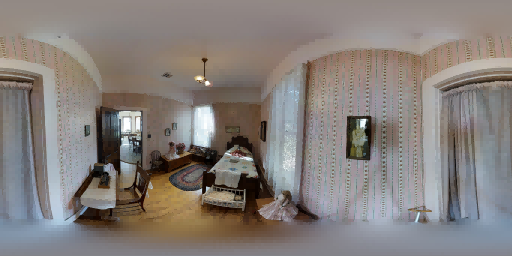}
	\includegraphics[width=.19\linewidth]{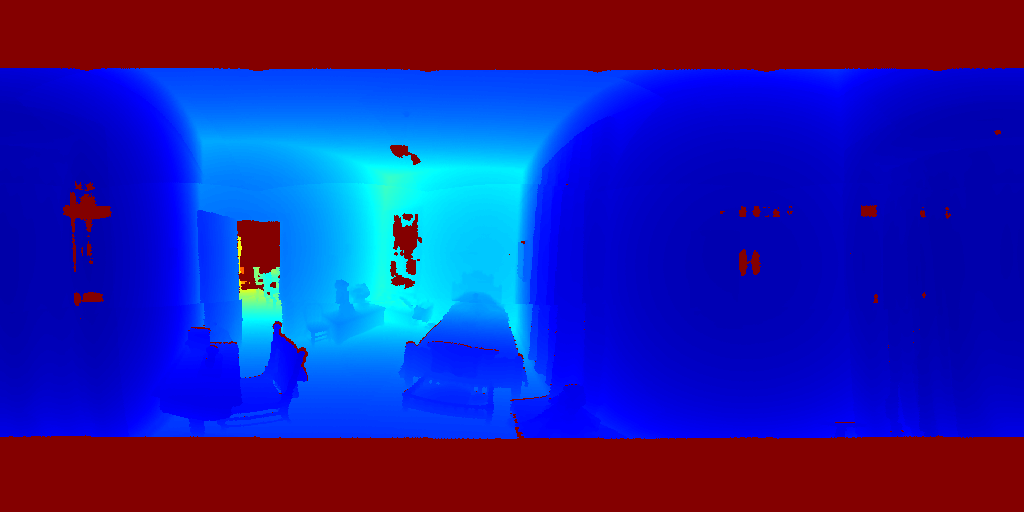}
	\includegraphics[width=.19\linewidth]{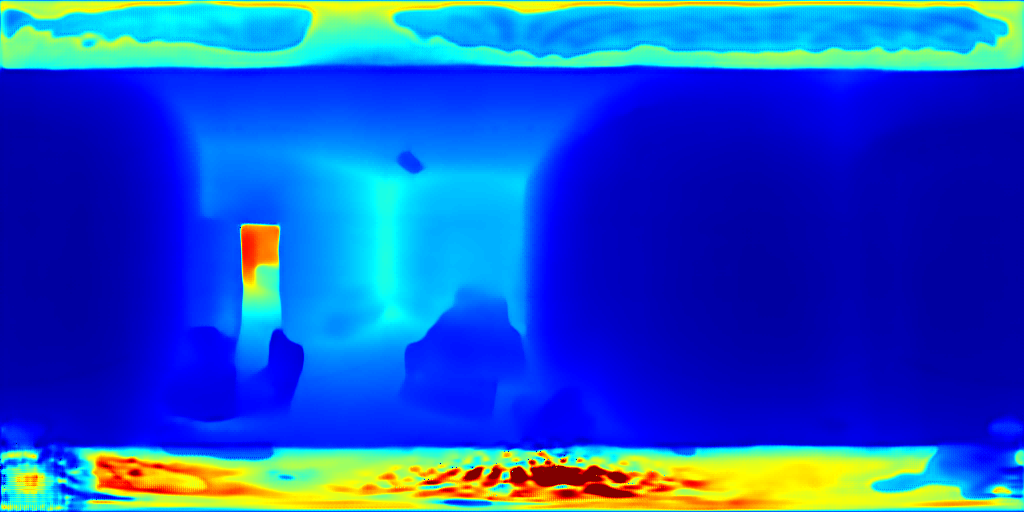}
	\includegraphics[width=.19\linewidth]{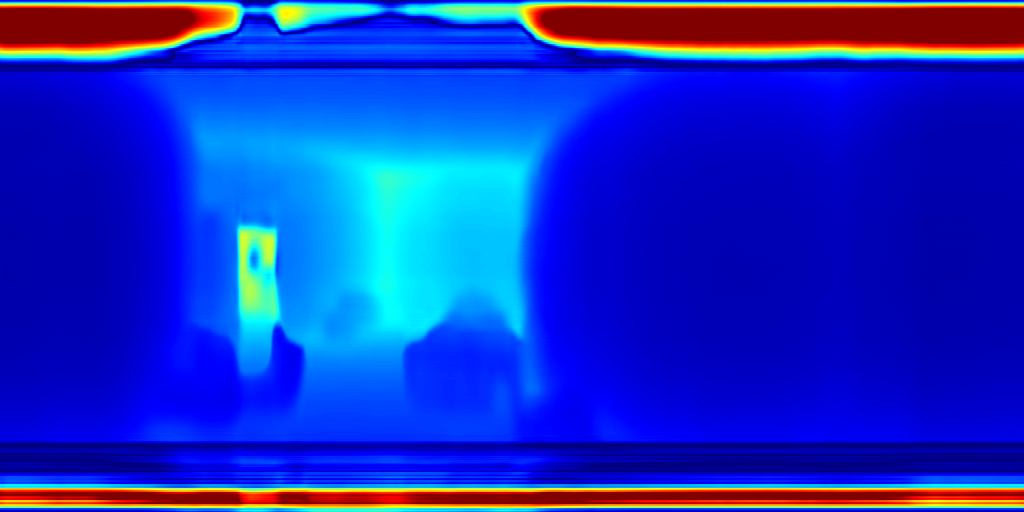}
	\includegraphics[width=.19\linewidth]{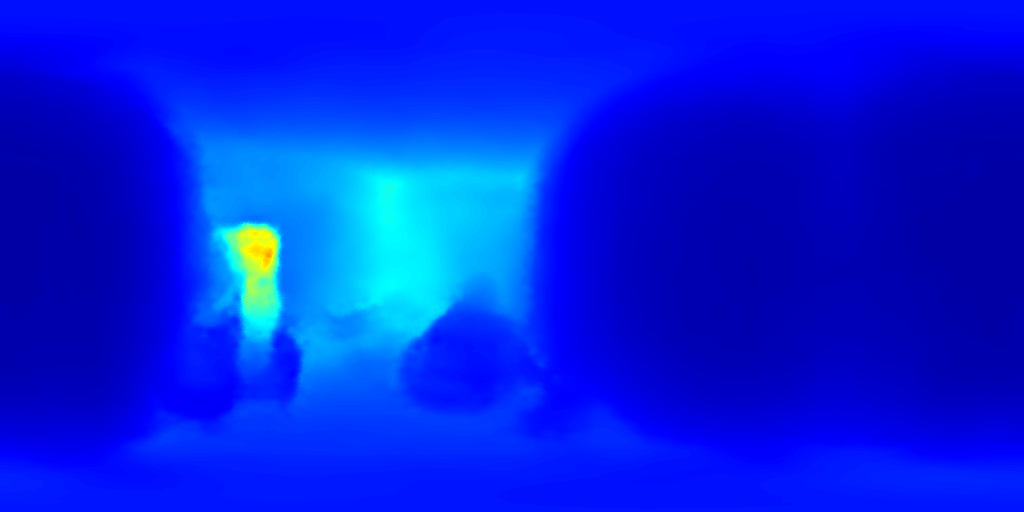}
	
	\caption{
		\textbf{Results of Matterport3D}
		Invalid parts of the depth map are set to red. BiFuse \cite{wang2020bifuse} and SliceNet \cite{pintore2021slicenet} cannot handle pole regions and generate lots of noise. Although there is no GT in the pole regions, SphereDepth still uses information from other regions to generate relatively reliable depths.
	}
	\label{fig:vis_mat3d}
\end{figure*}

\begin{figure*}
	\centering
	
	\hspace{0.01\linewidth} BiFuse \cite{wang2020bifuse}  \hspace{0.2\linewidth} SliceNet \cite{pintore2021slicenet} \hspace{0.22\linewidth} Ours
	
	\includegraphics[width=.3\linewidth]{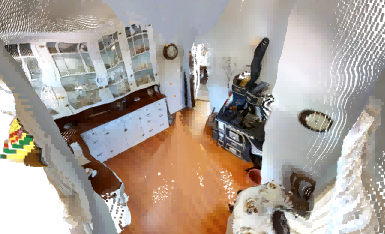}
	\includegraphics[width=.3\linewidth]{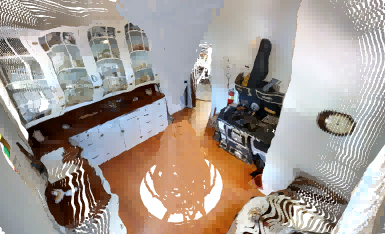}
	\includegraphics[width=.3\linewidth]{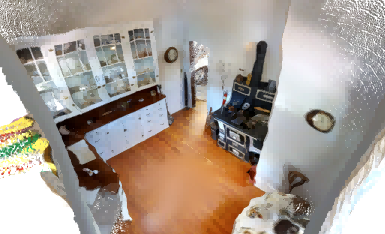}
	
	\includegraphics[width=.3\linewidth]{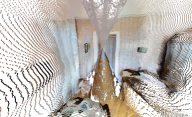}
	\includegraphics[width=.3\linewidth]{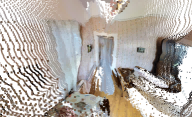}
	\includegraphics[width=.3\linewidth]{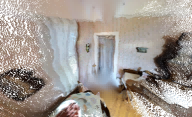}

	\caption{
		\textbf{Point cloud results on Matterport3D} SphereDepth produces a more complete and reliable 3D point cloud. However, there is a lot of noise in the point cloud of BiFuse \cite{wang2020bifuse} and SliceNet \cite{pintore2021slicenet}, which is challenging to use.
	}
	\label{fig:vis_mat3d_pointcloud}
\end{figure*}

\paragraph{MR and TR}

Our last ablation experiment discusses the impact of different spherical resolution strategies on SphereDepth. For a chosen SR, there are different strategies to set the MR and TR resolutions that lead to different GPU memory requirements, computing efficiency, and depth map quality.
The IR of 360D is 256$\times$512 and their corresponding $SR=7$ according to Table \ref{tab:look_up_resolution}. We test three different SR strategies, including $\{MR=5, TR=1\}$, $\{MR=5, TR=2\}$, and $\{MR=6, TR=1\}$. Table \ref{tab:choose_of_resolution} shows the results of different strategies. The SR strategy $\{MR=5, TR=1\}$, which is inconsistent with $SR=7$, gets the worse results. The results of  $\{MR=6, TR=1\}$ are the best, but it requires more GPU memory and more time during training for an epoch.
We should ensure $IR=SR$ and fully consider the computational efficiency in setting a larger MR based on these results. We choose $\{MR=5, TR=2\}$ for 360D, and $\{MR=6, TR=2\}$ for Stanford2D3D and Matterport3D.

\subsection{Quantitative Evaluation}

Following the results of ablation studies, we use the UNet as the encoder with our convolution kernel and the Log-loss function. Table \ref{tab:sota_results} shows the quantitative results of different methods on three datasets.

SphereDepth outperforms BiFuse \cite{wang2020bifuse} and achieves comparable results with SliceNet \cite{pintore2021slicenet} and HohoNet \cite{sun2021hohonet}. On the Standford2D3D dataset, SphereDepth does not achieve the best results, as the size of the dataset is small, and our method cannot benefit from the pre-trained model of ImageNet \cite{deng2009imagenet}. 
On the Matterport3D, SphereDepth almost achieves the best results among those existing studies with only a slight drop in some metrics. On the 360D, the metrics of SphereDepth are generally close to SliceNet but perform better than BiFuse. One interesting phenomenon is that SphereDepth does better in the $\delta$ metric than others almost on all three datasets, which shows SphereDepth has better-reconstructed results in the structure.

\begin{table*}
	\begin{center}
		\caption{Evaluation on three datasets. `S2D3D' is short for Standard2D3D. `M3D' is short for Matterport3D. }
		\label{tab:sota_results}
		\begin{tabular}{llccccccc}
			\hline\noalign{\smallskip}
			Dataset & Method & MRE$\downarrow$ & MAE$\downarrow$ & RMSE$\downarrow$ & RMSE(log)$\downarrow$ & $\delta_1\uparrow$ & $\delta_2\uparrow$ & $\delta_3\uparrow$ \\
			\noalign{\smallskip}
			\hline
			\noalign{\smallskip}
			\multirow{5}{*}{S2D3D}
			& FCRN \cite{laina2016deeper} & 0.1837 & 0.3428 & 0.5774 & 0.1100 & 0.7230 & 0.9207 & 0.9731\\ 
			& OmniDepth \cite{zioulis2018omnidepth} & 0.1996 & 0.3743 & 0.6152 & 0.1212 & 0.6877 & 0.8891 & 0.9578\\
			& BiFuse  \cite{wang2020bifuse}  & 0.1209 & 0.2343 & 0.4142 & 0.0787 & 0.8660 & 0.9580 & 0.9860\\ 
			& SliceNet\protect\footnotemark  \cite{pintore2021slicenet} & \textbf{0.0998} & \textbf{0.1737} & \textbf{0.3728} & 0.0765 & \textbf{0.9038} & 0.9623 & 0.9843 \\
			& HoHoNet \cite{sun2021hohonet} & 0.1014 & 0.2027 & 0.3834 & \textbf{0.0668} & 0.9054 & \textbf{0.9693} & \textbf{0.9886} \\
			& Ours & 0.1158 & 0.2323 & 0.4512 & 0.0754 & 0.8666 & 0.9642 & 0.9863 \\
			
			\hline
			
			\multirow{5}{*}{M3D} 
			& FCRN  \cite{laina2016deeper}    & 0.2409 & 0.4008 & 0.6704 & 0.1244 & 0.7703 & 0.9174 & 0.9617\\
			& OmniDepth \cite{zioulis2018omnidepth} & 0.2901 & 0.4838 & 0.7643 & 0.1450 & 0.6830 & 0.8794 & 0.9429\\
			& BiFuse \cite{wang2020bifuse}  & 0.2048 & 0.3470 & 0.6259 & 0.1134 & 0.8452 & 0.9319 & 0.9632\\
			& SliceNet \cite{pintore2021slicenet} & 0.1764 & 0.3296 & 0.6133 & 0.1045 & 0.8716 & 0.9483 & 0.9716 \\
			& HoHoNet \cite{sun2021hohonet} & 0.1488 & \textbf{0.2862} & \textbf{0.5138} & 0.0871 & \textbf{0.8786} & 0.9519 & \textbf{0.9771} \\
			& Ours & \textbf{0.1205} & 0.3311 & 0.5922 & \textbf{0.0806} & 0.8620 & \textbf{0.9519} & 0.9770 \\
			
			\hline
			
			\multirow{5}{*}{360D\protect\footnotemark}
			& FCRN \cite{laina2016deeper} & 0.0699 & 0.1381 & 0.2833 & 0.0473 & 0.9532 & 0.9905 & 0.9966\\
			& OmniDepth \cite{zioulis2018omnidepth} & 0.0931 & 0.1706 & 0.3171 & 0.0725 & 0.9092 & 0.9702 & 0.9851\\
			& BiFuse \cite{wang2020bifuse}  & 0.0615 & 0.1143 & 0.2440 & 0.0428 & 0.9699 & 0.9927 & 0.9969\\ 
			& SliceNet \cite{pintore2021slicenet} & \textbf{0.0467} & \textbf{0.1134} & \textbf{0.1323} & \textbf{0.0212} & \textbf{0.9788} & \textbf{0.9952} & 0.9969 \\ 
			& Ours & 0.0550 & 0.1145 & 0.2364 & 0.0369 & 0.9743 & 0.9944 & \textbf{0.9978} \\
			
			\hline
			
		\end{tabular}
	\end{center}
	\scriptsize{$^1$We recalculated all metrics using open source models.}
	\scriptsize{$^2$HoHoNet does not provide results on 360D.}
\end{table*}

\subsection{Qualitative Evaluation}

To further prove SphereDepth's reliability, we visualize some depth maps predicted by different methods. However, BiFuse and SliceNet only open-sourced parts of trained models; hence we cannot visualize them on all the datasets. We also compared the point clouds, which is much more straightforward to show the advantages of SphereDepth, instead of only comparing the depth map.

Fig. \ref{fig:vis_3d60} and Fig. \ref{fig:vis_2d3d} show the depth maps on 360D and Standford2D3D, respectively. SphereDepth achieves better results on the scene's structure but lacks details such as boundaries compared with the depth map. The second row in Fig. \ref{fig:vis_2d3d} shows that SphereDepth can even generate correct depth in GT missing regions, but SliceNet fails to do that. Fig. \ref{fig:vis_2d3d_pointcloud} shows two point clouds generated by SliceNet and SphereDepth. SphereDepth can generate a smoother and cleaner point cloud compared with SliceNet, which produces lots of noises in the point cloud.

Fig. \ref{fig:vis_mat3d} compares the depth maps generated by BiFuse, SliceNet and SphereDepth on Matterport3D. BiFuse and SliceNet cannot correctly estimate the depth of the ground and ceiling areas caused by lacking the corresponding GT in these regions. However, SphereDepth can still obtain relatively correct depths in these areas. The ground and ceiling areas are relatively small regions on the spherical domain instead of large areas in the panorama image in the equirectangular projection. Therefore, SphereDepth can automatically fill these missing regions with the predictions on the sphere space. Fig. \ref{fig:vis_mat3d_pointcloud}  further reflects the superiority of SphereDepth, which can generate complete and noise-removed point clouds.

\subsection{Limitations}

We propose SphereDepth for panorama depth estimation and achieved comparable results with the state-of-the-art results. However, SphereDepth still has some limitations. On the one hand, it is difficult for SphereDepth to process ultra-high-resolution panorama images because the higher SR will significantly increase the computational complexity. On the other hand, the network structure of SphereDepth is not tailor-made for the panorama image, and we still need more studies to find out the best.

\section{Conclusion}

This paper proposes a new depth estimation network for the panorama image, called SphereDepth, which can avoid the issues of distortion and discontinuities caused by projection methods, such as equirectangular projection or cube map projection. SphereDepth uses a customized convolution kernel to directly extract features on the spherical domain and perform depth estimation. The experimental results on the three datasets show that SphereDepth can obtain high-quality point clouds, which also shows that the SphereDepth can be further applied to other tasks under panorama images.

\clearpage
\newpage
\clearpage
\begin{center}
	{\Large \bf Supplementary material}
\end{center}
\setcounter{section}{0}

\section{Subdivision}

Subdivision is one of the key techniques of our method, which is used in triangle resolution (TR) and UnPooling. We demonstrate how Subdivision works by applying a subdivision to a triangle. Given a triangle with three points $(a,b,c)$, the subdivision operation can generate four triangles by Eq. \eqref{eq:tri}.

For the TR, subdivision helps us sample more triangle points to approximate mesh resolution (MR). As for UnPooling, the subdivision works in the decoder part of the network and splits a triangle into four triangles to generate a high-resolution depth map.

\begin{align} \label{eq:tri}
	\begin{cases}
		Triangle 0: (a,\frac{a+b}{2},\frac{a+c}{2}) \\
		Triangle 1: (b,\frac{a+b}{2},\frac{b+c}{2}) \\
		Triangle 2: (c,\frac{b+c}{2},\frac{a+c}{2}) \\
		Triangle 3: (\frac{b+c}{2},\frac{a+b}{2},\frac{a+c}{2}) \\
	\end{cases}
\end{align}

\section{Network Structure}

In this section, we introduce more details about the network structure. The network only contains an encoder and a decoder, which is much simpler than others\cite{wang2020bifuse,pintore2021slicenet}.

Before using the network to estimate the depth map, we have to convert the panorama image into the spherical domain by the spherical mesh with spherical resolution (SR) $S=M+T$, where $M$ refers to MR, and $T$ refers to TR. As the fundamental element of the spherical mesh is the triangle, we directly stack RGB values of all points generated by TR together and regrades them as features of the corresponding triangle, which means the network predicts the depth values of all points in each triangle. Based on the definition of MR and TR, the input shape of the network is $(1,20\times4^{M},c\times4^{T})$ and the output shape of the network is $(1,20\times4^{M},4^{T})$, in which $c$ represents the input feature number and $c=3$ when inputting an RGB panorama image.

Table \ref{tab:network} shows the parameter setting of SphereDepth, which is straightforward to implement.

\begin{table*}
	\centering
	\begin{tabular}{l|llll}
		
		\hline\noalign{\smallskip}
		Network Part & Name & Input Shape & Input Layer & Output Shape \\
		
		\hline
		
		\multirow{21}{*}{Encoder}
		
		&convb\_00 & (B,$20 \times 4^M$,$c \times 4^T$) & $pano$    & (B,$20 \times 4^M$, 64)   \\
		&convb\_01 & (B,$20 \times 4^M$, 64)      & convb\_00 & (B,$20 \times 4^M$, 64)   \\
		&pool\_0   & (B,$20 \times 4^M$, 64)     & convb\_01 & (B,$20 \times 4^{M-1}$, 64) \\
		
		&convb\_10 & (B,$20 \times 4^{M-1}$, 64)      & pool\_0   & (B,$20 \times 4^{M-1}$, 64)   \\
		&convb\_11 & (B,$20 \times 4^{M-1}$, 64)     & convb\_10 & (B,$20 \times 4^{M-1}$, 128)   \\
		&pool\_1   & (B,$20 \times 4^{M-1}$, 128)     & convb\_11 & (B,$20 \times 4^{M-2}$, 128)   \\	
		
		&convb\_20 & (B,$20 \times 4^{M-2}$, 128)     & pool\_1       & (B,$20 \times 4^{M-2}$, 128)   \\
		&convb\_21 & (B,$20 \times 4^{M-2}$, 128)     & convb\_20 & (B,$20 \times 4^{M-2}$, 128)   \\
		&convb\_22 & (B,$20 \times 4^{M-2}$, 128)     & convb\_21 & (B,$20 \times 4^{M-2}$, 256)   \\
		&pool\_2       & (B,$20 \times 4^{M-2}$, 256)      & convb\_22 & (B,$20 \times 4^{M-3}$, 256)   \\	
		
		&convb\_30 & (B,$20 \times 4^{M-3}$, 256)     & pool\_2       & (B,$20 \times 4^{M-3}$, 256)   \\
		&convb\_31 & (B,$20 \times 4^{M-3}$, 256)     & convb\_30 & (B,$20 \times 4^{M-3}$, 256)   \\
		&convb\_32 & (B,$20 \times 4^{M-3}$, 256)     & convb\_31 & (B,$20 \times 4^{M-3}$, 512)   \\
		&pool\_3       & (B,$20 \times 4^{M-3}$, 512)     & convb\_32 & (B,$20 \times 4^{M-4}$, 512)   \\	
		
		&convb\_40 & (B,$20 \times 4^{M-4}$, 512)     & pool\_3       & (B,$20 \times 4^{M-4}$, 512)   \\
		&convb\_41 & (B,$20 \times 4^{M-4}$, 512)     & convb\_40 & (B,$20 \times 4^{M-4}$, 512)   \\
		&convb\_42 & (B,$20 \times 4^{M-4}$, 512)     & convb\_41 & (B,$20 \times 4^{M-4}$, 512)   \\
		&pool\_4       & (B,$20 \times 4^{M-4}$, 512)      & conv\_b42 & (B,$20 \times 4^{M-5}$, 512)   \\			
		
		&convb\_50 & (B,$20 \times 4^{M-5}$, 512)     & pool\_4       & (B,$20 \times 4^{M-5}$, 512)   \\
		&convb\_51 & (B,$20 \times 4^{M-5}$, 512)     & convb\_50 & (B,$20 \times 4^{M-5}$, 512)   \\
		&convb\_52 & (B,$20 \times 4^{M-5}$, 512)     & convb\_51 & (B,$20 \times 4^{M-5}$, 512)   \\
		
		\hline
		
		\multirow{14}{*}{Decoder}
		&unpool\_4 & (B,$20 \times 4^{M-5}$, 512)     & convb\_52 & (B,$20 \times 4^{M-4}$, 512)  \\
		&dconv\_4 & (B,$20 \times 4^{M-4}$, 512)    & unpool\_4 & (B,$20 \times 4^{M-4}$, 512)  \\
		
		&unpool\_3  & (B,$20 \times 4^{M-4}$, 512)      & dconv\_4  & (B,$20 \times 4^{M-3}$, 512)  \\
		&dconv\_3  & (B,$20 \times 4^{M-3}$, 1024)    & unpool\_3,convb\_31 & (B,$20 \times 4^{M-3}$, 256) \\
		&output\_3  & (B,$20 \times 4^{M-3}$, 256 )  & dconv\_3  & (B,$20 \times 4^{M-3}$, $4^T$ ) \\
		
		&unpool\_2       & B,$20 \times 4^{M-3}$, 256)      & dconv\_3   & (B,$20 \times 4^{M-2}$, 256)  \\
		&dconv\_2  & (B,$20 \times 4^{M-2}$, 512)     & unpool\_2,convb\_21 & (B,$20 \times 4^{M-2}$, 128) \\
		&output\_2       & (B,$20 \times 4^{M-2}$, 256)   & dconv\_2  & (B,$20 \times 4^{M-2}$ ,$4^T$) \\
		
		&unpool\_1       & (B,$20 \times 4^{M-2}$, 128)      & dconv\_2   & (B,$20 \times 4^{M-1}$, 128)  \\
		&dconv\_1  & (B,$20 \times 4^{M-1}$, 128)      & unpool\_1,convb\_11 & (B,$20 \times 4^{M-1}$, 64) \\
		&output\_1       & (B,$20 \times 4^{M-1}$, 64)    & dconv\_1  & (B,$20 \times 4^{M-1}$ ,$4^T$) \\
		
		&unpool\_0       & (B,$20 \times 4^{M-1}$, 64)      & dconv\_1   & (B,$20 \times 4^{M}$, 64)  \\
		&dconvb\_0  & (B,$20 \times 4^{M}$, 128)      & unpool\_0,convb\_01 & (B,$20 \times 4^{M}$, 32) \\
		&output\_0       & (B,$20 \times 4^{M}$, 32)    & dconv\_0  & (B,$20 \times 4^{M}$, $4^T$)  \\
		
		\hline
		
	\end{tabular}

	\caption{
		\textbf{Structure of the Network}
		The layer $output_i(i=0,1,2,3)$ estimates depth map with different resolution. The deconv is a simple convolution layer with BatchNorm and ReLU. The convb has three convolution layers and a residual connection at the first layer and the last layer. $B$ represents the batch size, and $pano$ indicates the input. 
	}
	\label{tab:network}
\end{table*}

\section{Other Loss Functions}

In the ablation study, we compare the Log-loss with the Absolute-loss and the Huber-loss. Eq. \eqref{eq:loss_fun} shows the Absolute-loss and the Huber-loss used in the BiFuse \cite{wang2020bifuse} and SliceNet \cite{pintore2021slicenet}, in which $c$ determines where to switch from L1 to L2.

\begin{align} \label{eq:loss_fun}
	\begin{cases}
		abs\_loss &= |gt-pr|   \\
		huber\_loss &= 
		\begin{cases}
			| gt - pr | & |gt-pr|<c \\
			\frac{(gt-pr)^2+c^2}{2c} & |gt-pr|>=c
		\end{cases} \\
	\end{cases}
\end{align}

\section{Evaluation Metrics}

In this section, the indicators used for quantitative evaluation are presented. Following BiFuse \cite{wang2020bifuse} and SliceNet \cite{pintore2021slicenet}, we use five evaluation metrics, including MAE, MRE, RMSE, RMSE(log) and $\delta$. Eq. \eqref{eq:metrics} concludes how they are calculated.

During the evaluation, we ignore the pixels of which the ground truth depth is outside of the range $0.1 \sim 10$ meters for 360D \cite{zioulis2018omnidepth} and $0.1 \sim 16$ for Stanford2D3D \cite{armeni2017joint} and Matterport3D \cite{chang2017matterport3d} . We define $V$ as the set of valid pixels, and $N$ is the number of valid pixels in $V$. For each pixel $i$, we will use the following formulas to calculate the difference between the ground truth depth $gt_i$ and predict depth $pred_i$. The smaller is better for MAE, MRE, RMSE, and RMSE(log). For the $\delta$, the bigger is better.

\begin{align} \label{eq:metrics}
	\begin{cases}
		MAE = \sum_{i\in V}{|gt_i-pred_i|}   \\
		MRE = \sum_{i\in V}{\frac{|gt_i-pred_i|}{gt_i}} \\
		RMSE = \sqrt{\frac{\sum_{i\in V}(gt_i-pred_i)^2}{N}} \\
		RMSE_{log} = \sqrt{\frac{\sum_{i\in V}(log_{10}(gt_i)-log_{10}(pred_i))^2}{N}} \\
		\delta^n = \frac{\sum_{i\in V}max(\frac{gt_i}{pred_i},\frac{pred_i}{gt_i})<1.25^n}{N} \\
	\end{cases}
\end{align}

\section{Data Conversion}

\paragraph{Spherical Mesh and Equirectangular Projection}

This section introduces how to convert a panorama image in equirectangular projection to spherical mesh and vice versa.
For a panorama image with resolution $(W,H)$ in equirectangular projection, we define a 2d point on the  equirectangular image as $i=(u,v)$ and a 3d point on the spherical mesh as $p=(x,y,z)$.
We can project $p$ onto the equirectangular image by Eq. \eqref{eq:proj} and use the sampling method to gather the corresponding RGB value.

\begin{align}
	\label{eq:proj}
	\begin{cases}
		u = ( 1 + atan2(y,x)/\pi ) \times W/2 \\
		v = ( 0.5 + atan2(z,\sqrt{x^2+y^2})/\pi ) \times H
	\end{cases}
\end{align}

As for pixel $i$, we first calculate its corresponding point on the spherical mesh by Eq. \eqref{eq:to3d}, in which $\alpha$ and $\beta$ are the latitude and longitude, $(x_i,y_i,z_i)$ is the projected coordinates. Then, we can find its nearest triangle in the spherical mesh by KDTree \cite{bentley1975multidimensional} and assign the corresponding value to pixel $i$. 

\begin{align}
	\label{eq:to3d}
	\begin{cases}
		\alpha = 2*u/W -1  \\
		\beta = v/H - 0.5  \\
		x_i = cos(\beta)cos(\alpha) \\
		y_i = cos(\beta)sin(\alpha) \\
		z_i = sin(\beta)
	\end{cases}
\end{align}

\paragraph{Spherical Mesh To 3D Point Cloud}

Unlike perspective depth estimation, the depth value $d$ in the panorama image means the euclidean distance between the point $P=(X,Y,Z)$ in 3D real world and the panorama center. Therefore, we can convert a point $p=(x,y,z)$ on the spherical mesh to a 3D real world by Eq. \eqref{eq:point}.

\begin{align}
	\label{eq:point}
	\begin{cases}
		X = x \times d\\
		Y = y \times d\\ 
		Z = z \times d\\
	\end{cases}
\end{align}

\paragraph{Equirectangular Projection To 3D Point Cloud}

The panorama image in equirectangular projection can be converted into a point cloud by Eq. \eqref{eq:equ_proj}, following BiFuse \cite{wang2020bifuse}. 

\begin{align}
	\label{eq:equ_proj}
	\begin{cases}
		\alpha = 2*u/W -1  \\
		\beta = v/H - 0.5  \\
		X = cos(\beta)cos(\alpha) \times depth \\
		Y = cos(\beta)sin(\alpha) \times depth  \\
		Z = sin(\beta) \times depth \\
	\end{cases}
\end{align}

\section{Visualization and Discuss}

\begin{figure*}
	\centering
	
	\hspace{0.0\linewidth} RGB \hspace{0.18\linewidth}  GT \hspace{0.16\linewidth} Ours  \hspace{0.15\linewidth} PointCloud
	
	\includegraphics[width=.22\linewidth]{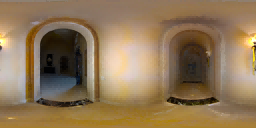}
	\includegraphics[width=.22\linewidth]{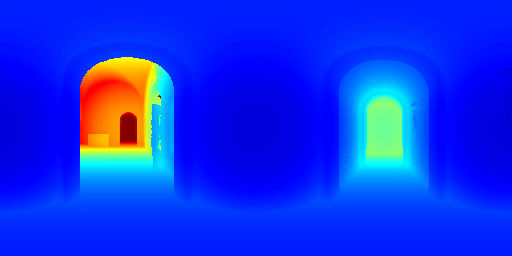}
	\includegraphics[width=.22\linewidth]{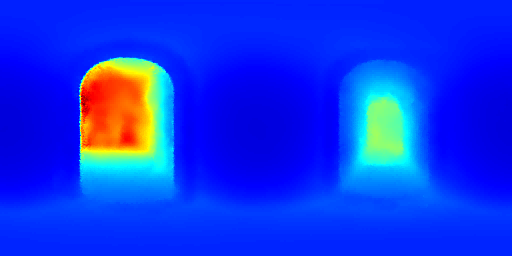}
	\includegraphics[width=.22\linewidth]{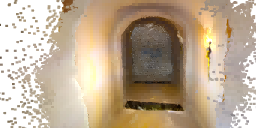}
	
	\includegraphics[width=.22\linewidth]{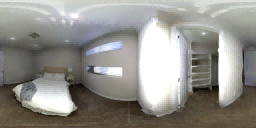}
	\includegraphics[width=.22\linewidth]{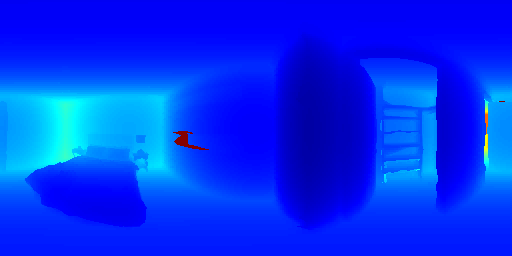}
	\includegraphics[width=.22\linewidth]{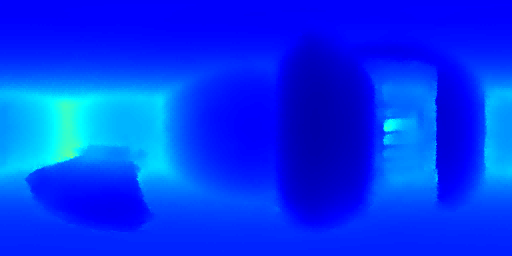}
	\includegraphics[width=.22\linewidth]{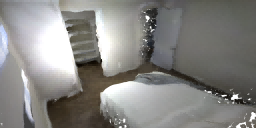}
	
	\includegraphics[width=.22\linewidth]{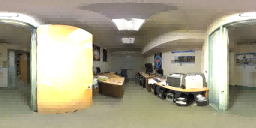}
	\includegraphics[width=.22\linewidth]{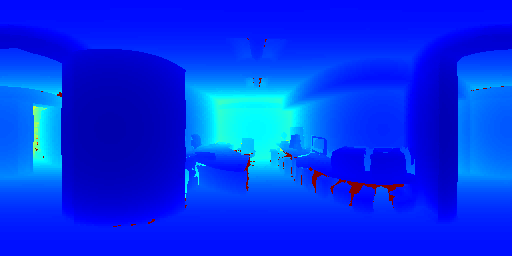}
	\includegraphics[width=.22\linewidth]{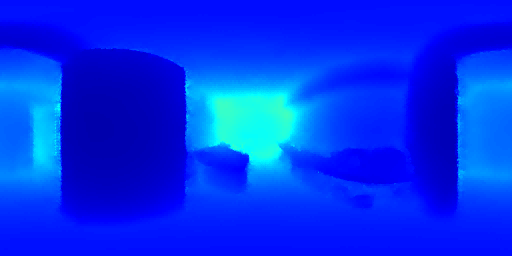}
	\includegraphics[width=.22\linewidth]{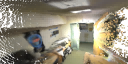}
	
	\includegraphics[width=.22\linewidth]{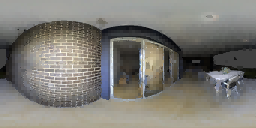}
	\includegraphics[width=.22\linewidth]{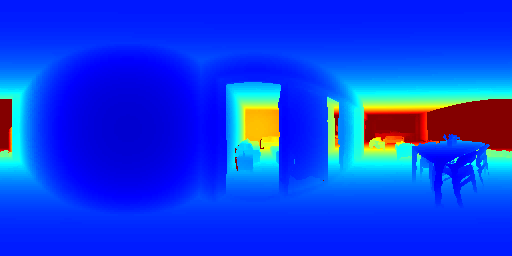}
	\includegraphics[width=.22\linewidth]{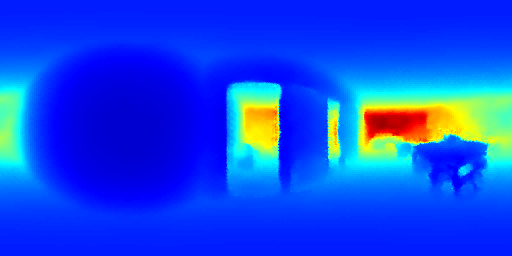}
	\includegraphics[width=.22\linewidth]{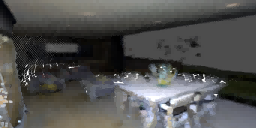}
	
	\includegraphics[width=.22\linewidth]{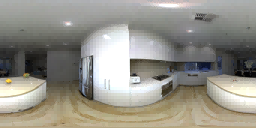}
	\includegraphics[width=.22\linewidth]{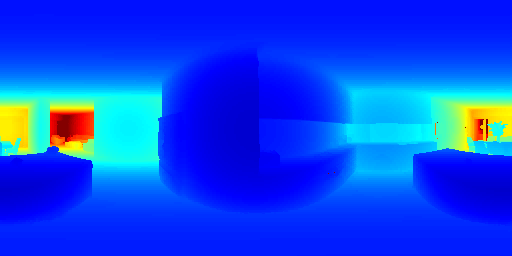}
	\includegraphics[width=.22\linewidth]{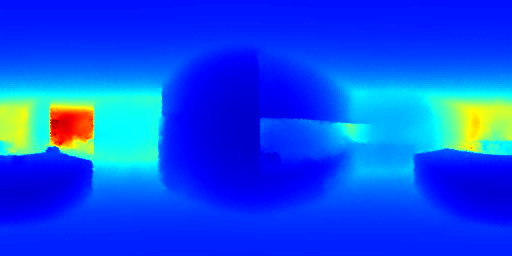}
	\includegraphics[width=.22\linewidth]{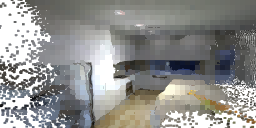}
	
	\includegraphics[width=.22\linewidth]{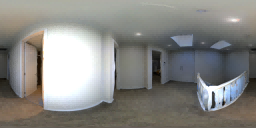}
	\includegraphics[width=.22\linewidth]{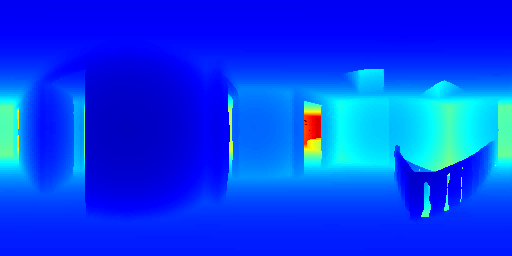}
	\includegraphics[width=.22\linewidth]{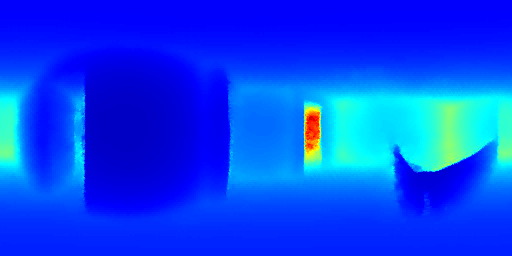}
	\includegraphics[width=.22\linewidth]{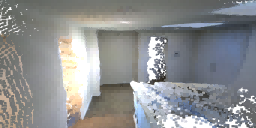}
	
	\includegraphics[width=.22\linewidth]{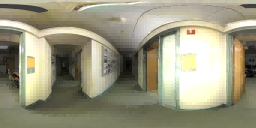}
	\includegraphics[width=.22\linewidth]{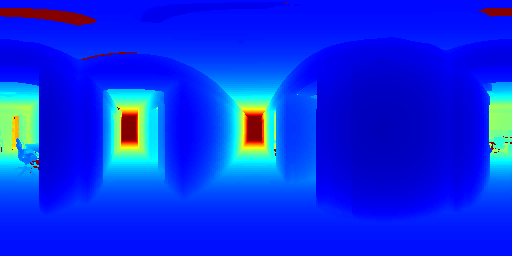}
	\includegraphics[width=.22\linewidth]{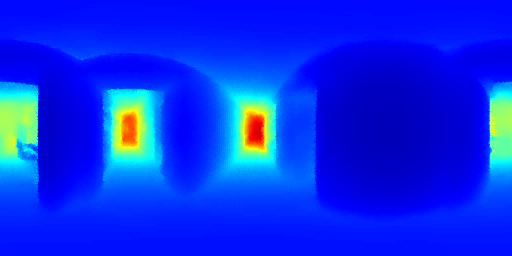}
	\includegraphics[width=.22\linewidth]{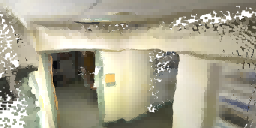}
	
	\includegraphics[width=.22\linewidth]{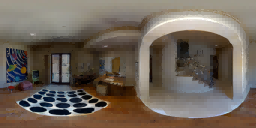}
	\includegraphics[width=.22\linewidth]{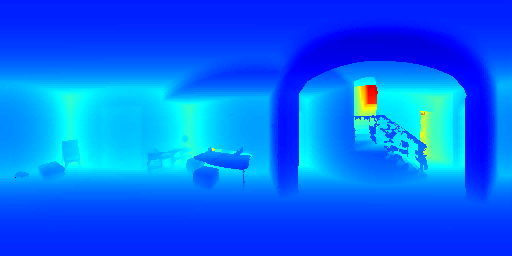}
	\includegraphics[width=.22\linewidth]{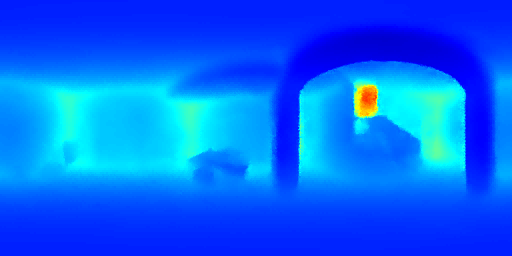}
	\includegraphics[width=.22\linewidth]{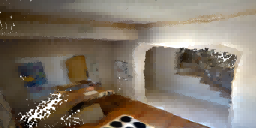}
	
	\includegraphics[width=.22\linewidth]{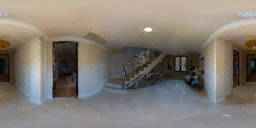}
	\includegraphics[width=.22\linewidth]{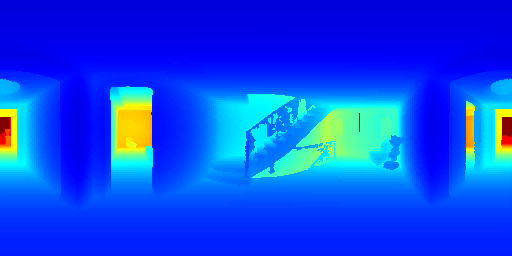}
	\includegraphics[width=.22\linewidth]{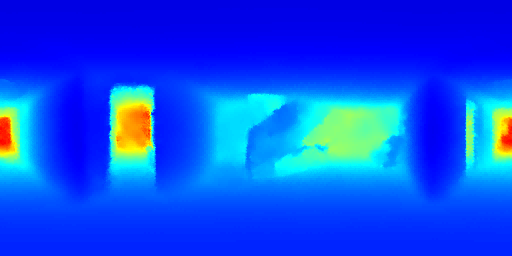}
	\includegraphics[width=.22\linewidth]{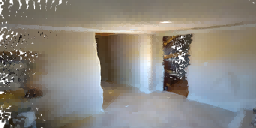}
	
	\includegraphics[width=.22\linewidth]{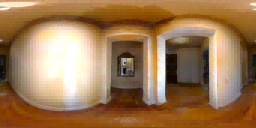}
	\includegraphics[width=.22\linewidth]{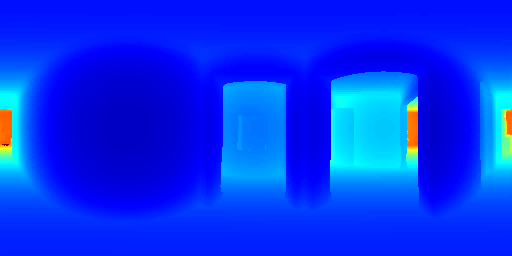}
	\includegraphics[width=.22\linewidth]{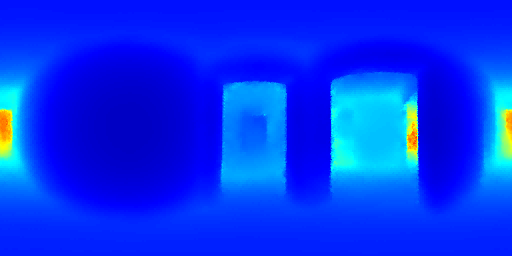}
	\includegraphics[width=.22\linewidth]{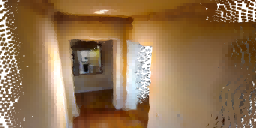}
	
	\includegraphics[width=.22\linewidth]{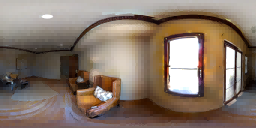}
	\includegraphics[width=.22\linewidth]{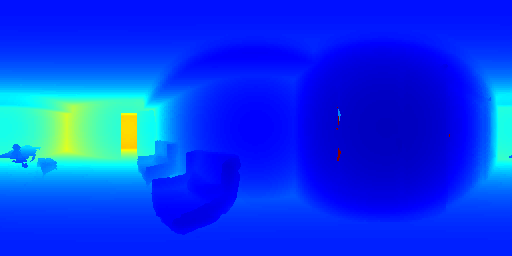}
	\includegraphics[width=.22\linewidth]{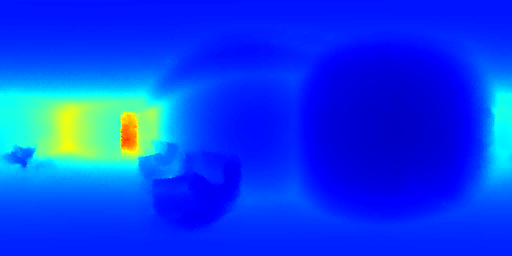}
	\includegraphics[width=.22\linewidth]{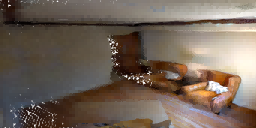}
	
	\caption{
		\textbf{Results of 360D}
		Invalid parts of the depth map are set to red.
	}
	
	\label{fig:more_360d}
\end{figure*}

\begin{figure*}
	\centering
	
	\hspace{0.0\linewidth} RGB \hspace{0.15\linewidth}  GT \hspace{0.15\linewidth} SliceNet  \cite{pintore2021slicenet} 
	\hspace{0.12\linewidth} Ours 
	
	\texttt{a}
	\includegraphics[width=.22\linewidth]{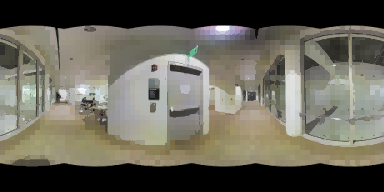}
	\includegraphics[width=.22\linewidth]{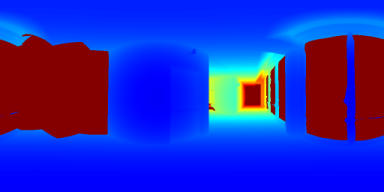}
	\includegraphics[width=.22\linewidth]{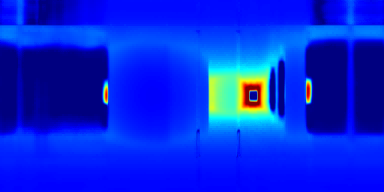}
	\includegraphics[width=.22\linewidth]{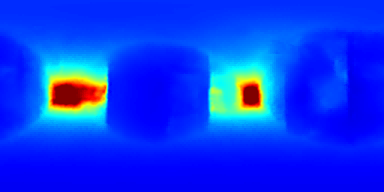}
	
	\texttt{b}
	\includegraphics[width=.22\linewidth]{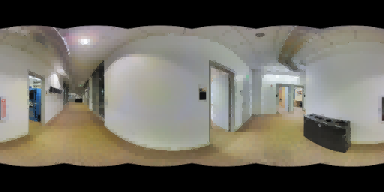}
	\includegraphics[width=.22\linewidth]{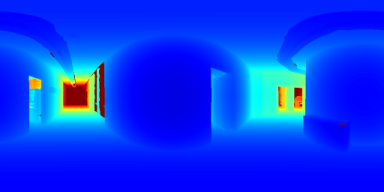}
	\includegraphics[width=.22\linewidth]{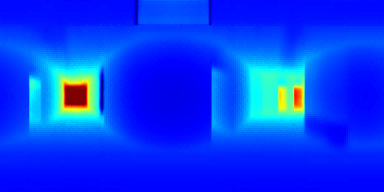}
	\includegraphics[width=.22\linewidth]{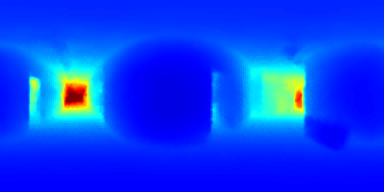}
	
	\texttt{c}
	\includegraphics[width=.22\linewidth]{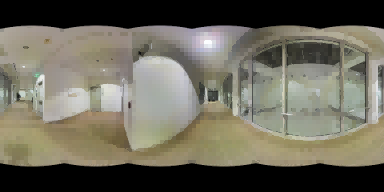}
	\includegraphics[width=.22\linewidth]{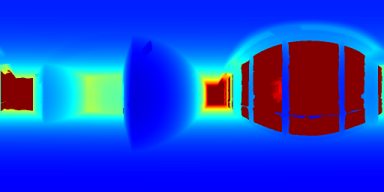}
	\includegraphics[width=.22\linewidth]{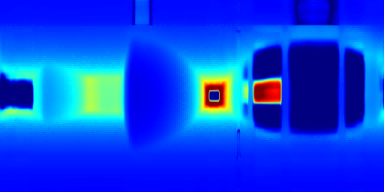}
	\includegraphics[width=.22\linewidth]{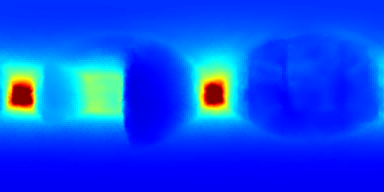}
	
	\texttt{d}
	\includegraphics[width=.22\linewidth]{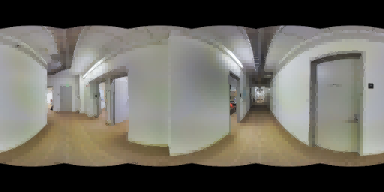}
	\includegraphics[width=.22\linewidth]{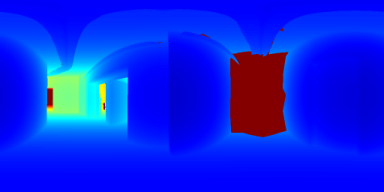}
	\includegraphics[width=.22\linewidth]{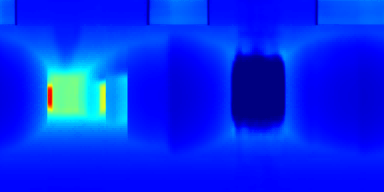}
	\includegraphics[width=.22\linewidth]{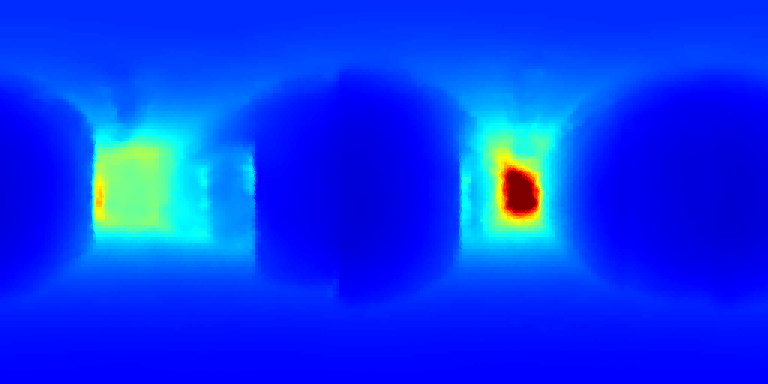}
	
	\texttt{e}
	\includegraphics[width=.22\linewidth]{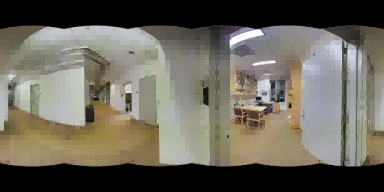}
	\includegraphics[width=.22\linewidth]{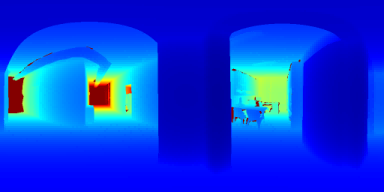}
	\includegraphics[width=.22\linewidth]{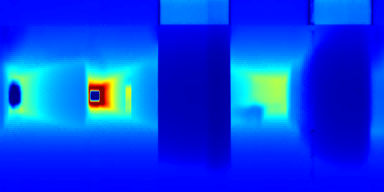}
	\includegraphics[width=.22\linewidth]{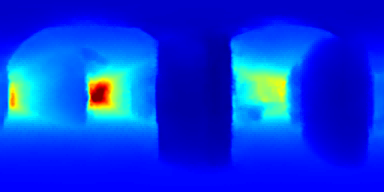}
	
	\texttt{f}
	\includegraphics[width=.22\linewidth]{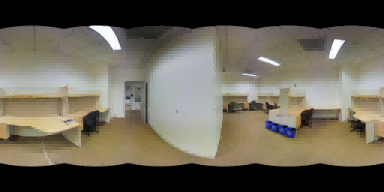}
	\includegraphics[width=.22\linewidth]{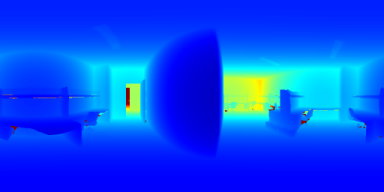}
	\includegraphics[width=.22\linewidth]{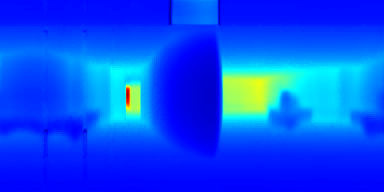}
	\includegraphics[width=.22\linewidth]{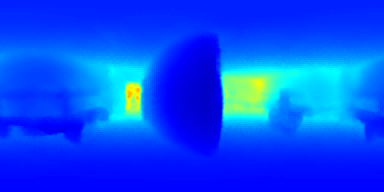}
	
	\texttt{g}
	\includegraphics[width=.22\linewidth]{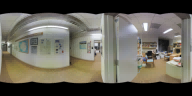}
	\includegraphics[width=.22\linewidth]{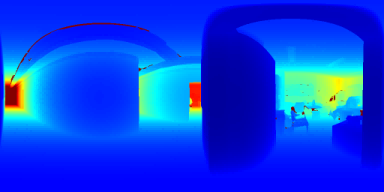}
	\includegraphics[width=.22\linewidth]{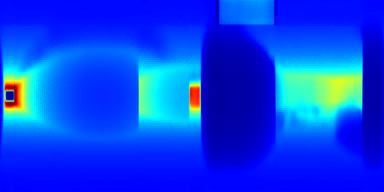}
	\includegraphics[width=.22\linewidth]{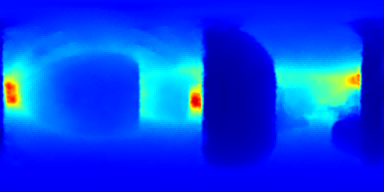}
	
	\texttt{h}
	\includegraphics[width=.22\linewidth]{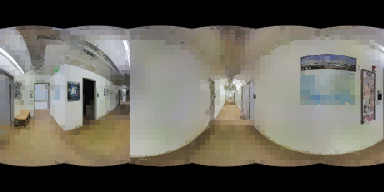}
	\includegraphics[width=.22\linewidth]{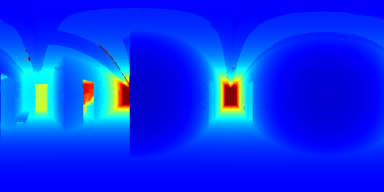}
	\includegraphics[width=.22\linewidth]{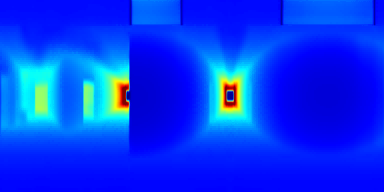}
	\includegraphics[width=.22\linewidth]{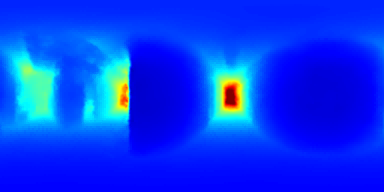}
	
	\texttt{i}
	\includegraphics[width=.22\linewidth]{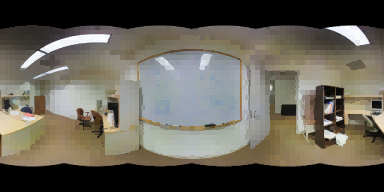}
	\includegraphics[width=.22\linewidth]{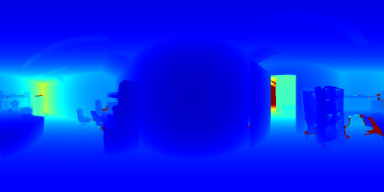}
	\includegraphics[width=.22\linewidth]{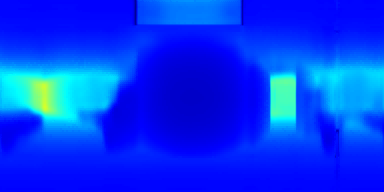}
	\includegraphics[width=.22\linewidth]{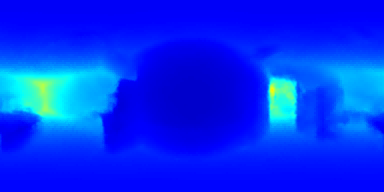}
	
	\texttt{j}
	\includegraphics[width=.22\linewidth]{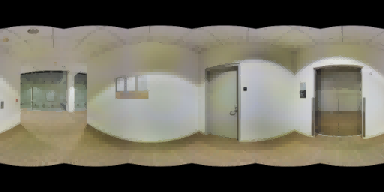}
	\includegraphics[width=.22\linewidth]{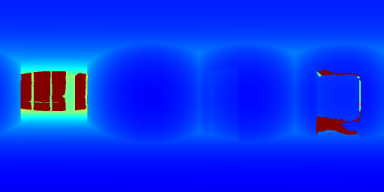}
	\includegraphics[width=.22\linewidth]{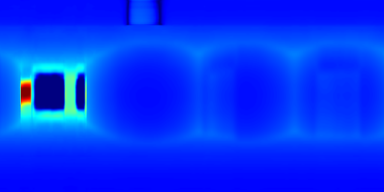}
	\includegraphics[width=.22\linewidth]{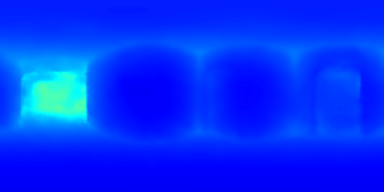}
	
	\texttt{k}
	\includegraphics[width=.22\linewidth]{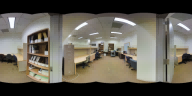}
	\includegraphics[width=.22\linewidth]{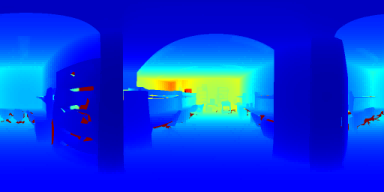}
	\includegraphics[width=.22\linewidth]{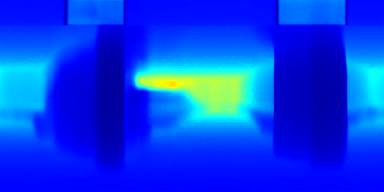}
	\includegraphics[width=.22\linewidth]{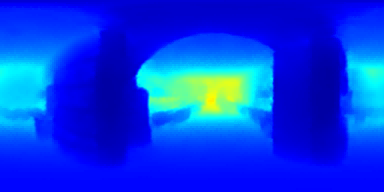}
	
	\caption{
		\textbf{Results of Standford2D3D}
		Invalid parts of the depth map are set to red.
	}
	\label{fig:more_2d3d}
\end{figure*}

\begin{figure*}
	\centering
	
	\hspace{0.0\linewidth} SliceNet \cite{pintore2021slicenet} 
	\hspace{0.25\linewidth} Ours 
	
	\texttt{a}
	\includegraphics[width=.35\linewidth]{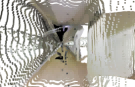}
	\includegraphics[width=.35\linewidth]{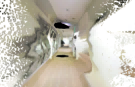}
	
	\texttt{b}
	\includegraphics[width=.35\linewidth]{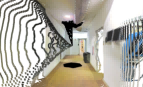}
	\includegraphics[width=.35\linewidth]{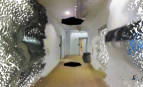}
	
	\texttt{c}
	\includegraphics[width=.35\linewidth]{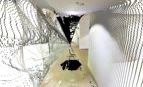}
	\includegraphics[width=.35\linewidth]{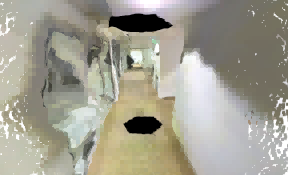}
	
	\texttt{d}
	\includegraphics[width=.35\linewidth]{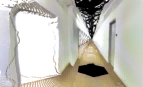}
	\includegraphics[width=.35\linewidth]{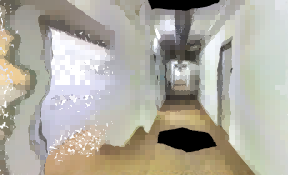}
	
	\texttt{e}
	\includegraphics[width=.35\linewidth]{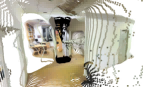}
	\includegraphics[width=.35\linewidth]{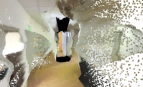}

	\caption{
		\textbf{PointCloud Results of Standford2D3D  from \texttt{a} to \texttt{e} }
	}
	
	\label{fig:more_2d3d_pointcloud_a_e}
\end{figure*}

\begin{figure*}
	\centering
	
	\hspace{0.0\linewidth} SliceNet \cite{pintore2021slicenet} 
	\hspace{0.25\linewidth} Ours 
	
	\texttt{f}
	\includegraphics[width=.35\linewidth]{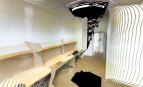}
	\includegraphics[width=.35\linewidth]{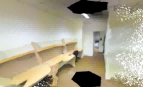}
	
	\texttt{g}
	\includegraphics[width=.35\linewidth]{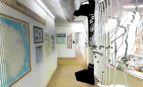}
	\includegraphics[width=.35\linewidth]{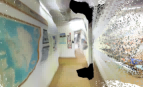}
	
	\texttt{h}
	\includegraphics[width=.35\linewidth]{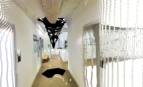}
	\includegraphics[width=.35\linewidth]{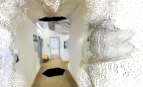}
	
	\texttt{i}
	\includegraphics[width=.35\linewidth]{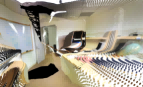}
	\includegraphics[width=.35\linewidth]{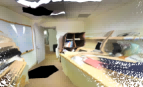}
	
	\texttt{j}
	\includegraphics[width=.35\linewidth]{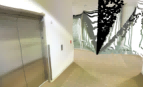}
	\includegraphics[width=.35\linewidth]{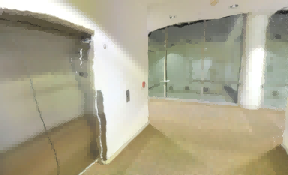}
	
	\texttt{k}
	\includegraphics[width=.35\linewidth]{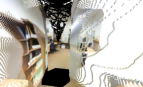}
	\includegraphics[width=.35\linewidth]{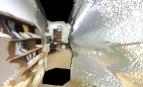}
	
	\caption{
		\textbf{PointCloud Results of Standford2D3D  from \texttt{f} to \texttt{k}}
	}
	
	\label{fig:more_2d3d_pointcloud_f_k}
\end{figure*}

\begin{figure*}
	\centering
	
	\hspace{0.0\linewidth} RGB \hspace{0.12\linewidth}  GT \hspace{0.09\linewidth} BiFuse \cite{wang2020bifuse} \hspace{0.06\linewidth} SliceNet \cite{pintore2021slicenet} \hspace{0.09\linewidth} Ours 
	
	\texttt{a}
	\includegraphics[width=.16\linewidth]{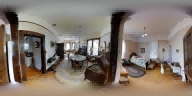}
	\includegraphics[width=.16\linewidth]{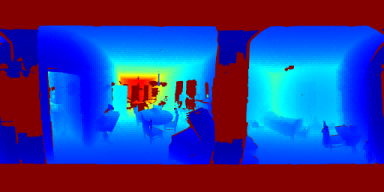}
	\includegraphics[width=.16\linewidth]{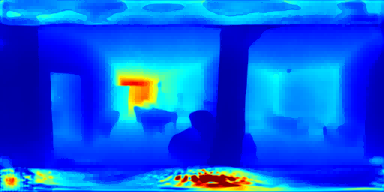}
	\includegraphics[width=.16\linewidth]{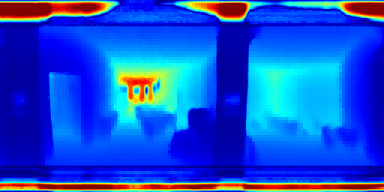}
	\includegraphics[width=.16\linewidth]{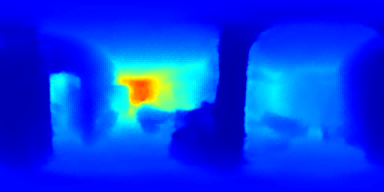}
	
	\texttt{b}
	\includegraphics[width=.16\linewidth]{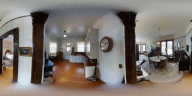}
	\includegraphics[width=.16\linewidth]{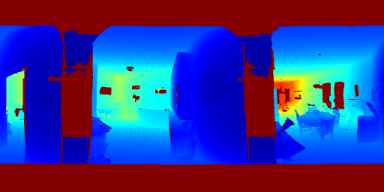}
	\includegraphics[width=.16\linewidth]{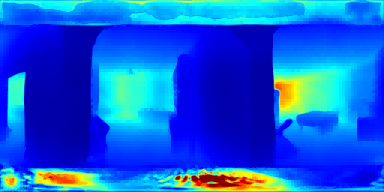}
	\includegraphics[width=.16\linewidth]{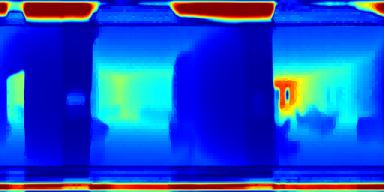}
	\includegraphics[width=.16\linewidth]{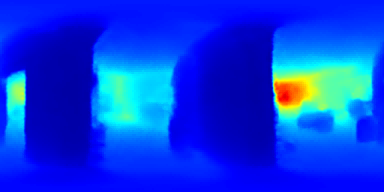}
	
	\texttt{c}
	\includegraphics[width=.16\linewidth]{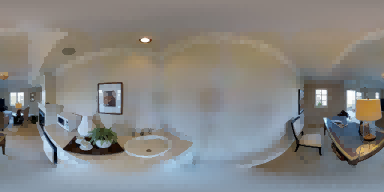}
	\includegraphics[width=.16\linewidth]{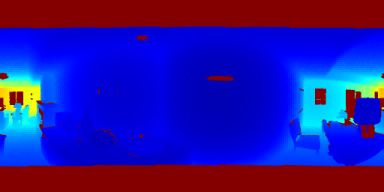}
	\includegraphics[width=.16\linewidth]{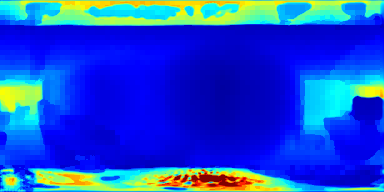}
	\includegraphics[width=.16\linewidth]{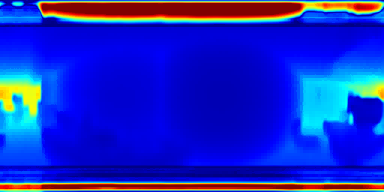}
	\includegraphics[width=.16\linewidth]{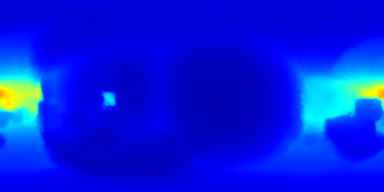}
	
	\texttt{d}
	\includegraphics[width=.16\linewidth]{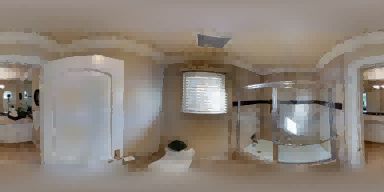}
	\includegraphics[width=.16\linewidth]{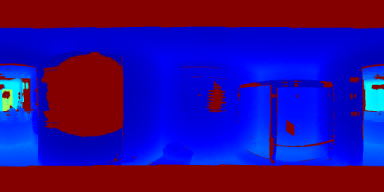}
	\includegraphics[width=.16\linewidth]{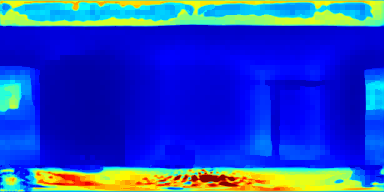}
	\includegraphics[width=.16\linewidth]{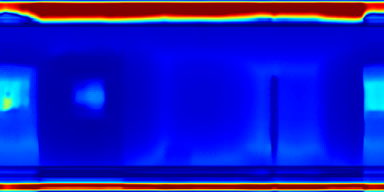}
	\includegraphics[width=.16\linewidth]{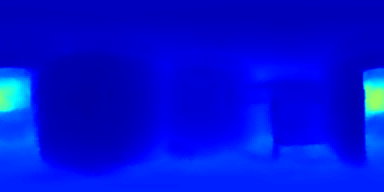}
	
	\texttt{e}
	\includegraphics[width=.16\linewidth]{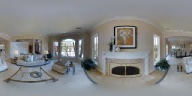}
	\includegraphics[width=.16\linewidth]{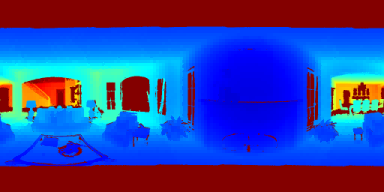}
	\includegraphics[width=.16\linewidth]{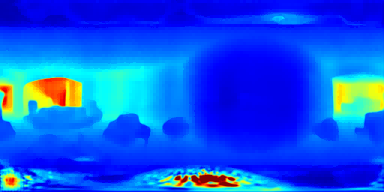}
	\includegraphics[width=.16\linewidth]{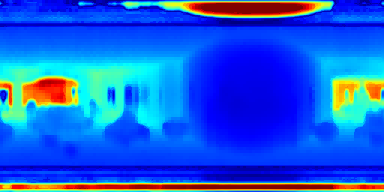}
	\includegraphics[width=.16\linewidth]{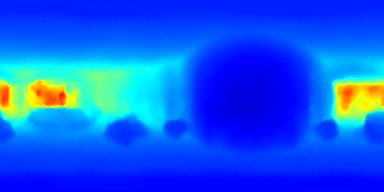}
	
	\texttt{f}
	\includegraphics[width=.16\linewidth]{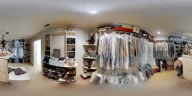}
	\includegraphics[width=.16\linewidth]{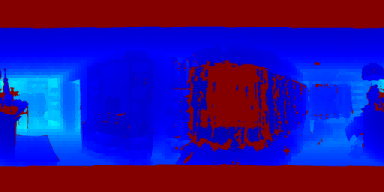}
	\includegraphics[width=.16\linewidth]{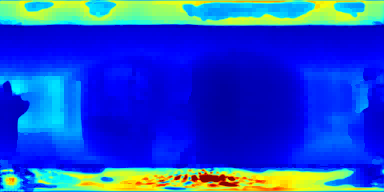}
	\includegraphics[width=.16\linewidth]{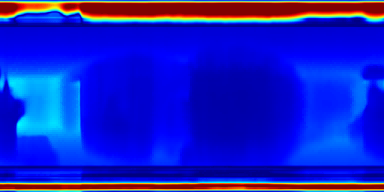}
	\includegraphics[width=.16\linewidth]{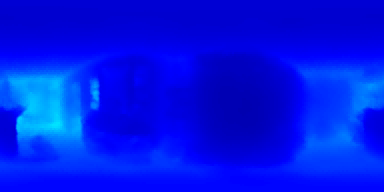}
	
	\texttt{g}
	\includegraphics[width=.16\linewidth]{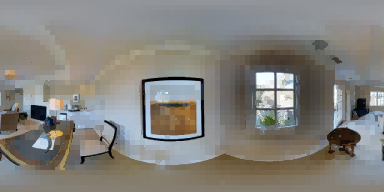}
	\includegraphics[width=.16\linewidth]{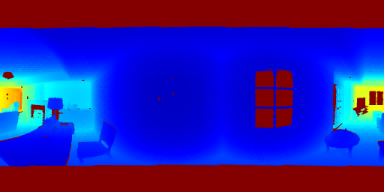}
	\includegraphics[width=.16\linewidth]{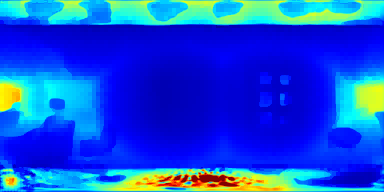}
	\includegraphics[width=.16\linewidth]{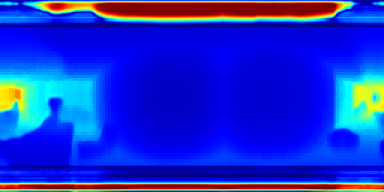}
	\includegraphics[width=.16\linewidth]{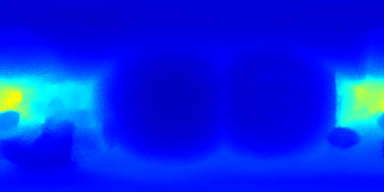}
	
	\texttt{h}
	\includegraphics[width=.16\linewidth]{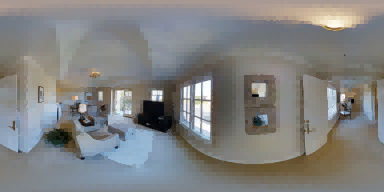}
	\includegraphics[width=.16\linewidth]{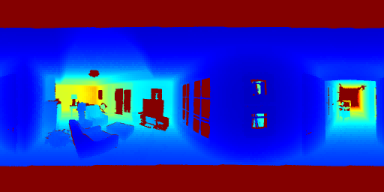}
	\includegraphics[width=.16\linewidth]{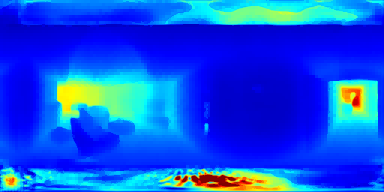}
	\includegraphics[width=.16\linewidth]{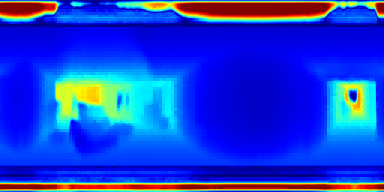}
	\includegraphics[width=.16\linewidth]{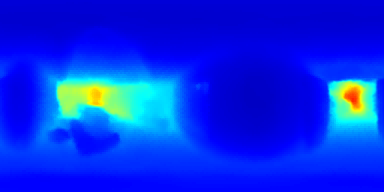}
	
	\texttt{i}
	\includegraphics[width=.16\linewidth]{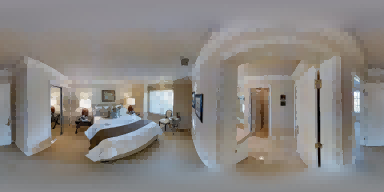}
	\includegraphics[width=.16\linewidth]{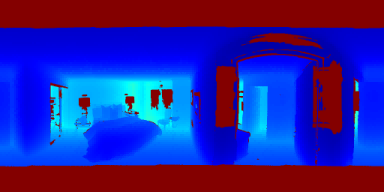}
	\includegraphics[width=.16\linewidth]{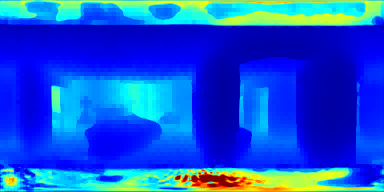}
	\includegraphics[width=.16\linewidth]{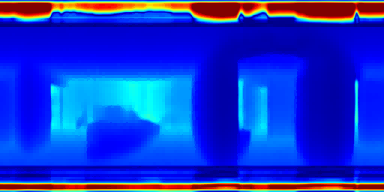}
	\includegraphics[width=.16\linewidth]{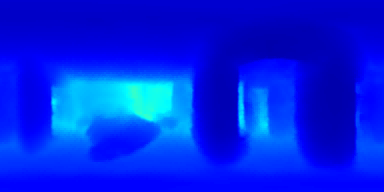}
	
	\texttt{j}
	\includegraphics[width=.16\linewidth]{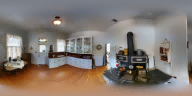}
	\includegraphics[width=.16\linewidth]{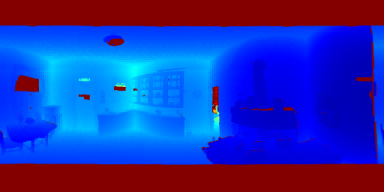}
	\includegraphics[width=.16\linewidth]{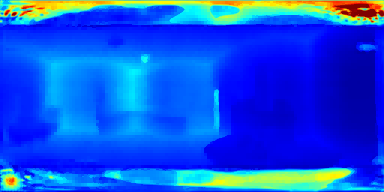}
	\includegraphics[width=.16\linewidth]{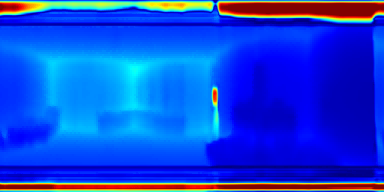}
	\includegraphics[width=.16\linewidth]{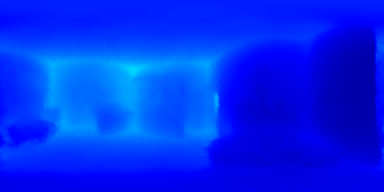}
	
	\texttt{k}
	\includegraphics[width=.16\linewidth]{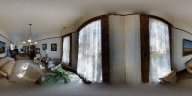}
	\includegraphics[width=.16\linewidth]{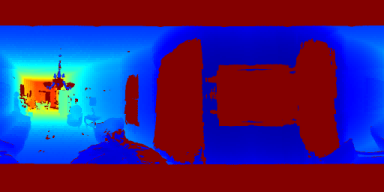}
	\includegraphics[width=.16\linewidth]{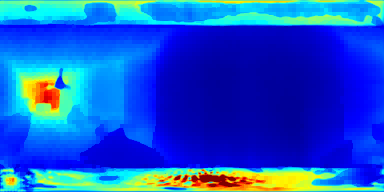}
	\includegraphics[width=.16\linewidth]{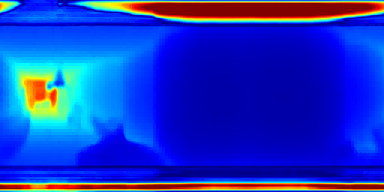}
	\includegraphics[width=.16\linewidth]{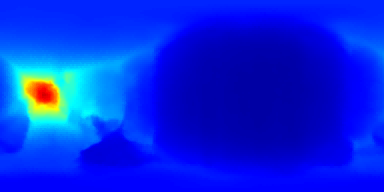}
	
	\texttt{l}
	\includegraphics[width=.16\linewidth]{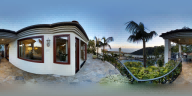}
	\includegraphics[width=.16\linewidth]{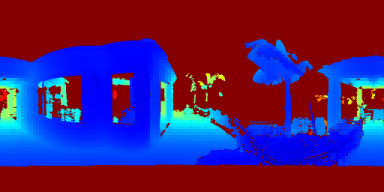}
	\includegraphics[width=.16\linewidth]{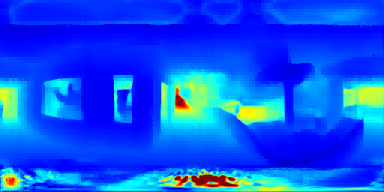}
	\includegraphics[width=.16\linewidth]{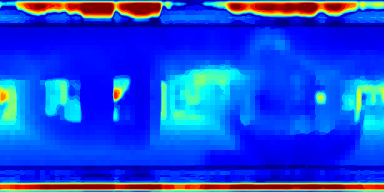}
	\includegraphics[width=.16\linewidth]{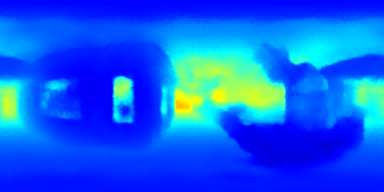}
	
	\texttt{m}
	\includegraphics[width=.16\linewidth]{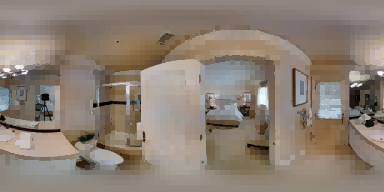}
	\includegraphics[width=.16\linewidth]{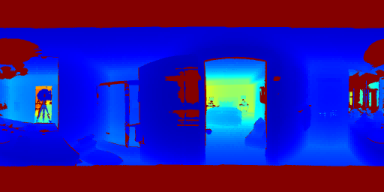}
	\includegraphics[width=.16\linewidth]{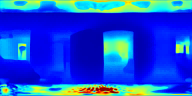}
	\includegraphics[width=.16\linewidth]{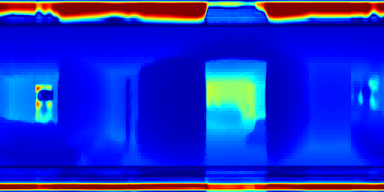}
	\includegraphics[width=.16\linewidth]{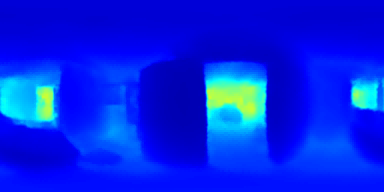}
	
	\texttt{n}
	\includegraphics[width=.16\linewidth]{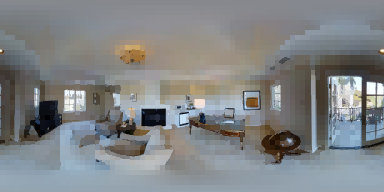}
	\includegraphics[width=.16\linewidth]{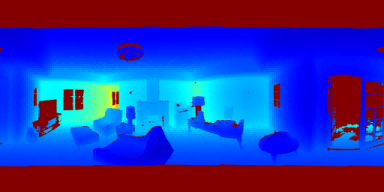}
	\includegraphics[width=.16\linewidth]{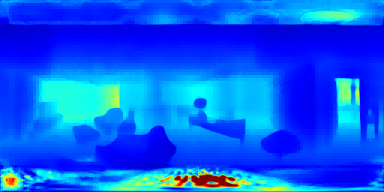}
	\includegraphics[width=.16\linewidth]{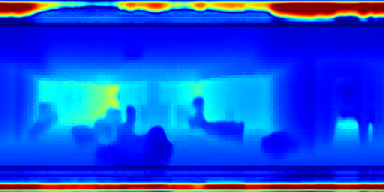}
	\includegraphics[width=.16\linewidth]{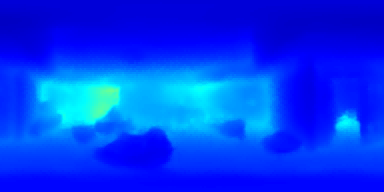}
	
	\texttt{o}
	\includegraphics[width=.16\linewidth]{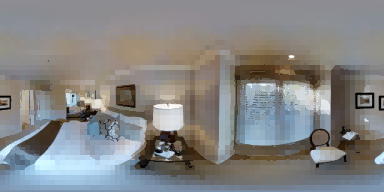}
	\includegraphics[width=.16\linewidth]{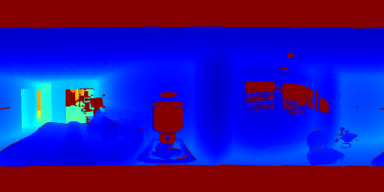}
	\includegraphics[width=.16\linewidth]{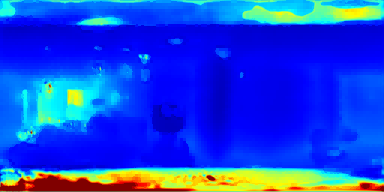}
	\includegraphics[width=.16\linewidth]{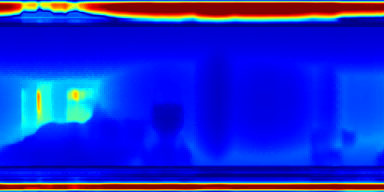}
	\includegraphics[width=.16\linewidth]{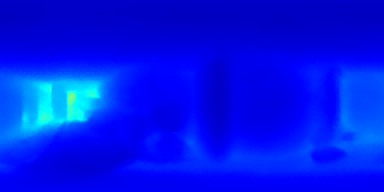}

	\caption{
		\textbf{Results of Matterport3D}
		Invalid parts of the depth map are set to red.
	}
	\label{fig:more_mat3d}
	
\end{figure*}

\begin{figure*}
	\centering
	
	\hspace{0.0\linewidth} BiFuse \cite{wang2020bifuse}  \hspace{0.15\linewidth} SliceNet \cite{pintore2021slicenet} \hspace{0.15\linewidth} Ours 
	
	\texttt{a}
	\includegraphics[width=.25\linewidth]{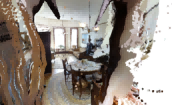}
	\includegraphics[width=.25\linewidth]{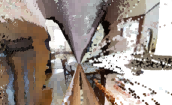}
	\includegraphics[width=.25\linewidth]{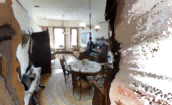}
	
	\texttt{b}
	\includegraphics[width=.25\linewidth]{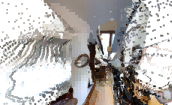}
	\includegraphics[width=.25\linewidth]{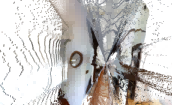}
	\includegraphics[width=.25\linewidth]{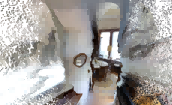}
	
	\texttt{c}
	\includegraphics[width=.25\linewidth]{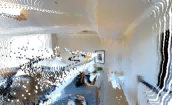}
	\includegraphics[width=.25\linewidth]{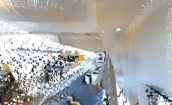}
	\includegraphics[width=.25\linewidth]{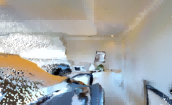}
	
	\texttt{d}
	\includegraphics[width=.25\linewidth]{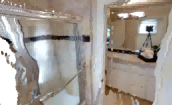}
	\includegraphics[width=.25\linewidth]{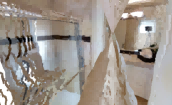}
	\includegraphics[width=.25\linewidth]{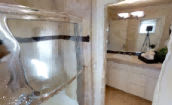}
	
	\texttt{e}
	\includegraphics[width=.25\linewidth]{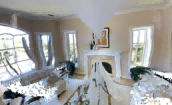}
	\includegraphics[width=.25\linewidth]{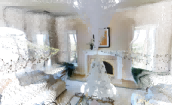}
	\includegraphics[width=.25\linewidth]{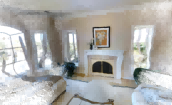}
	
	\texttt{f}
	\includegraphics[width=.25\linewidth]{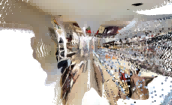}
	\includegraphics[width=.25\linewidth]{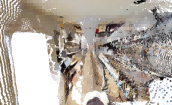}
	\includegraphics[width=.25\linewidth]{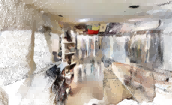}
	
	\texttt{g}
	\includegraphics[width=.25\linewidth]{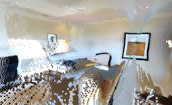}
	\includegraphics[width=.25\linewidth]{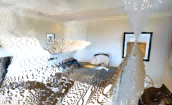}
	\includegraphics[width=.25\linewidth]{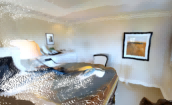}
	\caption{
		\textbf{PointCloud Results of Matterport3D from \texttt{a} to \texttt{g}}
	}
	\label{fig:more_mat3d_pointcloud_a_g}
\end{figure*}

\begin{figure*}
	\centering
	
	\hspace{0.0\linewidth} BiFuse \cite{wang2020bifuse}  \hspace{0.15\linewidth} SliceNet \cite{pintore2021slicenet} \hspace{0.15\linewidth} Ours 
	
	\texttt{h}
	\includegraphics[width=.25\linewidth]{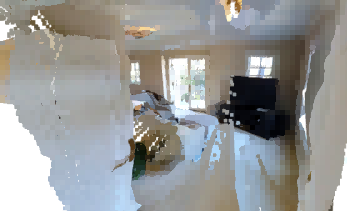}
	\includegraphics[width=.25\linewidth]{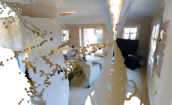}
	\includegraphics[width=.25\linewidth]{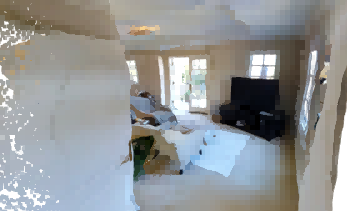}
	
	\texttt{i}
	\includegraphics[width=.25\linewidth]{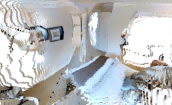}
	\includegraphics[width=.25\linewidth]{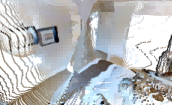}
	\includegraphics[width=.25\linewidth]{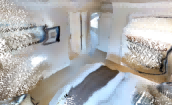}
	
	\texttt{j}
	\includegraphics[width=.25\linewidth]{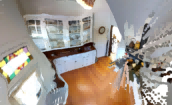}
	\includegraphics[width=.25\linewidth]{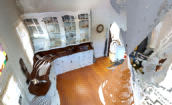}
	\includegraphics[width=.25\linewidth]{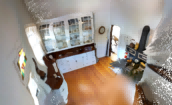}
	
	\texttt{k}
	\includegraphics[width=.25\linewidth]{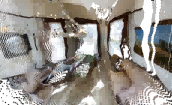}
	\includegraphics[width=.25\linewidth]{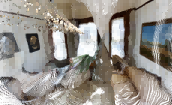}
	\includegraphics[width=.25\linewidth]{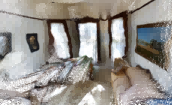}
	
	\texttt{l}
	\includegraphics[width=.25\linewidth]{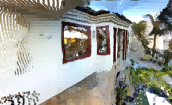}
	\includegraphics[width=.25\linewidth]{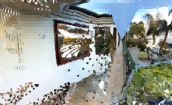}
	\includegraphics[width=.25\linewidth]{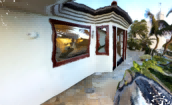}
	
	\texttt{m}
	\includegraphics[width=.25\linewidth]{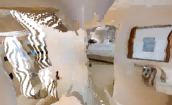}
	\includegraphics[width=.25\linewidth]{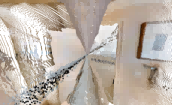}
	\includegraphics[width=.25\linewidth]{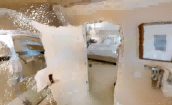}
	
	\texttt{n}
	\includegraphics[width=.25\linewidth]{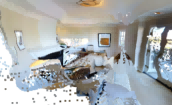}
	\includegraphics[width=.25\linewidth]{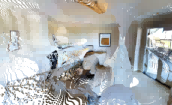}
	\includegraphics[width=.25\linewidth]{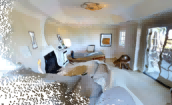}
	
	\texttt{o}
	\includegraphics[width=.25\linewidth]{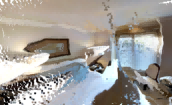}
	\includegraphics[width=.25\linewidth]{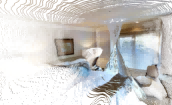}
	\includegraphics[width=.25\linewidth]{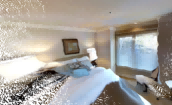}
	
	\caption{
		\textbf{PointCloud Results of Matterport3D from \texttt{h} to \texttt{o} }
	}
	\label{fig:more_mat3d_pointcloud_h_o}
\end{figure*}

We add more visualization results in this section to highlight the advantages of SphereDepth. Fig. \ref{fig:more_360d} shows ours results on 360D. Fig.\ref{fig:more_2d3d} shows depth map results of SliceNet and SphereDepth, and Fig. \ref{fig:more_2d3d_pointcloud_a_e} and Fig. \ref{fig:more_2d3d_pointcloud_f_k} show corresponding point clouds. Fig. \ref{fig:more_mat3d} shows depth map results of BiFuse, SliceNet and SphereDepth, and Fig. \ref{fig:more_mat3d_pointcloud_a_g} and Fig. \ref{fig:more_mat3d_pointcloud_h_o} show point clouds.

BiFuse and SliceNet suffer from discontinuity and distortion caused by projection and cannot estimate depth values of the ceiling and the floor. However, SphereDepth can overcome these disadvantages and
generate complete point clouds and recover 3d layouts of scenes.

\clearpage
{\small
	\bibliographystyle{ieee_fullname}
	\bibliography{egbib}

\begin{thebibliography}{10}\itemsep=-1pt

\bibitem{alhashim2018high}
Ibraheem Alhashim and Peter Wonka.
\newblock High quality monocular depth estimation via transfer learning.
\newblock {\em arXiv preprint arXiv:1812.11941}, 2018.

\bibitem{armeni2017joint}
Iro Armeni, Sasha Sax, Amir~R Zamir, and Silvio Savarese.
\newblock Joint 2d-3d-semantic data for indoor scene understanding.
\newblock {\em arXiv preprint arXiv:1702.01105}, 2017.

\bibitem{bentley1975multidimensional}
Jon~Louis Bentley.
\newblock Multidimensional binary search trees used for associative searching.
\newblock {\em Communications of the ACM}, 18(9):509--517, 1975.

\bibitem{bhat2021adabins}
Shariq~Farooq Bhat, Ibraheem Alhashim, and Peter Wonka.
\newblock Adabins: Depth estimation using adaptive bins.
\newblock In {\em Proceedings of the IEEE/CVF Conference on Computer Vision and
  Pattern Recognition}, pages 4009--4018, 2021.

\bibitem{bian2021unsupervised}
Jia-Wang Bian, Huangying Zhan, Naiyan Wang, Zhichao Li, Le Zhang, Chunhua Shen,
  Ming-Ming Cheng, and Ian Reid.
\newblock Unsupervised scale-consistent depth learning from video.
\newblock {\em International Journal of Computer Vision}, pages 1--17, 2021.

\bibitem{chang2017matterport3d}
Angel Chang, Angela Dai, Thomas Funkhouser, Maciej Halber, Matthias Niessner,
  Manolis Savva, Shuran Song, Andy Zeng, and Yinda Zhang.
\newblock Matterport3d: Learning from rgb-d data in indoor environments.
\newblock {\em arXiv preprint arXiv:1709.06158}, 2017.

\bibitem{cheng2018cube}
Hsien-Tzu Cheng, Chun-Hung Chao, Jin-Dong Dong, Hao-Kai Wen, Tyng-Luh Liu, and
  Min Sun.
\newblock Cube padding for weakly-supervised saliency prediction in 360 videos.
\newblock In {\em Proceedings of the IEEE Conference on Computer Vision and
  Pattern Recognition}, pages 1420--1429, 2018.

\bibitem{cohen2018spherical}
Taco~S Cohen, Mario Geiger, Jonas K{\"o}hler, and Max Welling.
\newblock Spherical cnns.
\newblock {\em arXiv preprint arXiv:1801.10130}, 2018.

\bibitem{coors2018spherenet}
Benjamin Coors, Alexandru~Paul Condurache, and Andreas Geiger.
\newblock Spherenet: Learning spherical representations for detection and
  classification in omnidirectional images.
\newblock In {\em Proceedings of the European Conference on Computer Vision
  (ECCV)}, pages 518--533, 2018.

\bibitem{deng2009imagenet}
Jia Deng, Wei Dong, Richard Socher, Li-Jia Li, Kai Li, and Li Fei-Fei.
\newblock Imagenet: A large-scale hierarchical image database.
\newblock In {\em 2009 IEEE conference on computer vision and pattern
  recognition}, pages 248--255. Ieee, 2009.

\bibitem{dosovitskiy2020image}
Alexey Dosovitskiy, Lucas Beyer, Alexander Kolesnikov, Dirk Weissenborn,
  Xiaohua Zhai, Thomas Unterthiner, Mostafa Dehghani, Matthias Minderer, Georg
  Heigold, Sylvain Gelly, et~al.
\newblock An image is worth 16x16 words: Transformers for image recognition at
  scale.
\newblock {\em arXiv preprint arXiv:2010.11929}, 2020.

\bibitem{eder2020tangent}
Marc Eder, Mykhailo Shvets, John Lim, and Jan-Michael Frahm.
\newblock Tangent images for mitigating spherical distortion.
\newblock In {\em Proceedings of the IEEE/CVF Conference on Computer Vision and
  Pattern Recognition}, pages 12426--12434, 2020.

\bibitem{eigen2014depth}
David Eigen, Christian Puhrsch, and Rob Fergus.
\newblock Depth map prediction from a single image using a multi-scale deep
  network.
\newblock In {\em NIPS}, 2014.

\bibitem{feng2019meshnet}
Yutong Feng, Yifan Feng, Haoxuan You, Xibin Zhao, and Yue Gao.
\newblock Meshnet: Mesh neural network for 3d shape representation.
\newblock In {\em Proceedings of the AAAI Conference on Artificial
  Intelligence}, volume~33, pages 8279--8286, 2019.

\bibitem{fernandez2020corners}
Clara Fernandez-Labrador, Jose~M Facil, Alejandro Perez-Yus, C{\'e}dric
  Demonceaux, Javier Civera, and Jose~J Guerrero.
\newblock Corners for layout: End-to-end layout recovery from 360 images.
\newblock {\em IEEE Robotics and Automation Letters}, 5(2):1255--1262, 2020.

\bibitem{fu2018deep}
Huan Fu, Mingming Gong, Chaohui Wang, Kayhan Batmanghelich, and Dacheng Tao.
\newblock Deep ordinal regression network for monocular depth estimation.
\newblock In {\em Proceedings of the IEEE conference on computer vision and
  pattern recognition}, pages 2002--2011, 2018.

\bibitem{monodepth2017}
Clement Godard, Oisin Mac~Aodha, and Gabriel~J. Brostow.
\newblock Unsupervised monocular depth estimation with left-right consistency.
\newblock In {\em Proceedings of the IEEE Conference on Computer Vision and
  Pattern Recognition (CVPR)}, July 2017.

\bibitem{godard2019digging}
Cl{\'e}ment Godard, Oisin Mac~Aodha, Michael Firman, and Gabriel~J Brostow.
\newblock Digging into self-supervised monocular depth estimation.
\newblock In {\em Proceedings of the IEEE/CVF International Conference on
  Computer Vision}, pages 3828--3838, 2019.

\bibitem{hanocka2019meshcnn}
Rana Hanocka, Amir Hertz, Noa Fish, Raja Giryes, Shachar Fleishman, and Daniel
  Cohen-Or.
\newblock Meshcnn: a network with an edge.
\newblock {\em ACM Transactions on Graphics (TOG)}, 38(4):1--12, 2019.

\bibitem{he2016deep}
Kaiming He, Xiangyu Zhang, Shaoqing Ren, and Jian Sun.
\newblock Deep residual learning for image recognition.
\newblock In {\em Proceedings of the IEEE conference on computer vision and
  pattern recognition}, pages 770--778, 2016.

\bibitem{hu2020jittor}
Shi-Min Hu, Dun Liang, Guo-Ye Yang, Guo-Wei Yang, and Wen-Yang Zhou.
\newblock Jittor: a novel deep learning framework with meta-operators and
  unified graph execution.
\newblock {\em Science China Information Sciences}, 63(222103):1--21, 2020.

\bibitem{hu2021subdivision}
Shi-Min Hu, Zheng-Ning Liu, Meng-Hao Guo, Jun-Xiong Cai, Jiahui Huang,
  Tai-Jiang Mu, and Ralph~R Martin.
\newblock Subdivision-based mesh convolution networks.
\newblock {\em ACM Transactions on Graphics (TOG)}, 41:1--16, 2022.

\bibitem{jiang2018spherical}
Chiyu~Max Jiang, Jingwei Huang, Karthik Kashinath, Philip Marcus, Matthias
  Niessner, et~al.
\newblock Spherical cnns on unstructured grids.
\newblock In {\em International Conference on Learning Representations}, 2018.

\bibitem{jiang2021unifuse}
Hualie Jiang, Zhe Sheng, Siyu Zhu, Zilong Dong, and Rui Huang.
\newblock Unifuse: Unidirectional fusion for 360 panorama depth estimation.
\newblock {\em IEEE Robotics and Automation Letters}, 6(2):1519--1526, 2021.

\bibitem{jin2020geometric}
Lei Jin, Yanyu Xu, Jia Zheng, Junfei Zhang, Rui Tang, Shugong Xu, Jingyi Yu,
  and Shenghua Gao.
\newblock Geometric structure based and regularized depth estimation from 360
  indoor imagery.
\newblock In {\em Proceedings of the IEEE/CVF Conference on Computer Vision and
  Pattern Recognition}, pages 889--898, 2020.

\bibitem{kingma2014adam}
Diederik~P Kingma and Jimmy Ba.
\newblock Adam: A method for stochastic optimization.
\newblock {\em arXiv preprint arXiv:1412.6980}, 2014.

\bibitem{laina2016deeper}
Iro Laina, Christian Rupprecht, Vasileios Belagiannis, Federico Tombari, and
  Nassir Navab.
\newblock Deeper depth prediction with fully convolutional residual networks.
\newblock In {\em 2016 Fourth international conference on 3D vision (3DV)},
  pages 239--248. IEEE, 2016.

\bibitem{lee2019spherephd}
Yeonkun Lee, Jaeseok Jeong, Jongseob Yun, Wonjune Cho, and Kuk-Jin Yoon.
\newblock Spherephd: Applying cnns on a spherical polyhedron representation of
  360deg images.
\newblock In {\em Proceedings of the IEEE/CVF Conference on Computer Vision and
  Pattern Recognition}, pages 9181--9189, 2019.

\bibitem{luo2020consistent}
Xuan Luo, Jia-Bin Huang, Richard Szeliski, Kevin Matzen, and Johannes Kopf.
\newblock Consistent video depth estimation.
\newblock {\em ACM Transactions on Graphics (TOG)}, 39(4):71--1, 2020.

\bibitem{miangoleh2021boosting}
S~Mahdi~H Miangoleh, Sebastian Dille, Long Mai, Sylvain Paris, and Yagiz Aksoy.
\newblock Boosting monocular depth estimation models to high-resolution via
  content-adaptive multi-resolution merging.
\newblock In {\em Proceedings of the IEEE/CVF Conference on Computer Vision and
  Pattern Recognition}, pages 9685--9694, 2021.

\bibitem{pintore2021slicenet}
Giovanni Pintore, Marco Agus, Eva Almansa, Jens Schneider, and Enrico Gobbetti.
\newblock Slicenet: deep dense depth estimation from a single indoor panorama
  using a slice-based representation.
\newblock In {\em Proceedings of the IEEE/CVF Conference on Computer Vision and
  Pattern Recognition}, pages 11536--11545, 2021.

\bibitem{pintore2021deep3dlayout}
Giovanni Pintore, Eva Almansa, Marco Agus, and Enrico Gobbetti.
\newblock Deep3dlayout: 3d reconstruction of an indoor layout from a spherical
  panoramic image.
\newblock {\em ACM Transactions on Graphics (TOG)}, 40(6):1--12, 2021.

\bibitem{ranftl2021vision}
Ren{\'e} Ranftl, Alexey Bochkovskiy, and Vladlen Koltun.
\newblock Vision transformers for dense prediction.
\newblock In {\em Proceedings of the IEEE/CVF International Conference on
  Computer Vision}, pages 12179--12188, 2021.

\bibitem{ranftl2020towards}
R Ranftl, K Lasinger, D Hafner, K Schindler, and V Koltun.
\newblock Towards robust monocular depth estimation: Mixing datasets for
  zero-shot cross-dataset transfer.
\newblock {\em IEEE Transactions on Pattern Analysis and Machine Intelligence},
  2020.

\bibitem{ronneberger2015u}
Olaf Ronneberger, Philipp Fischer, and Thomas Brox.
\newblock U-net: Convolutional networks for biomedical image segmentation.
\newblock In {\em International Conference on Medical image computing and
  computer-assisted intervention}, pages 234--241. Springer, 2015.

\bibitem{sfm2018}
Muhamad Risqi~U. Saputra, Andrew Markham, and Niki Trigoni.
\newblock Visual slam and structure from motion in dynamic environments: A
  survey.
\newblock {\em ACM Computing Surveys}, 51(2), Feb. 2018.

\bibitem{schonberger2016structure}
Johannes~L Schonberger and Jan-Michael Frahm.
\newblock Structure-from-motion revisited.
\newblock In {\em Proceedings of the IEEE conference on computer vision and
  pattern recognition}, pages 4104--4113, 2016.

\bibitem{su2017learning}
Yu-Chuan Su and Kristen Grauman.
\newblock Learning spherical convolution for fast features from 360 imagery.
\newblock {\em Advances in Neural Information Processing Systems}, 30:529--539,
  2017.

\bibitem{sun2019horizonnet}
Cheng Sun, Chi-Wei Hsiao, Min Sun, and Hwann-Tzong Chen.
\newblock Horizonnet: Learning room layout with 1d representation and pano
  stretch data augmentation.
\newblock In {\em Proceedings of the IEEE/CVF Conference on Computer Vision and
  Pattern Recognition}, pages 1047--1056, 2019.

\bibitem{sun2021hohonet}
Cheng Sun, Min Sun, and Hwann-Tzong Chen.
\newblock Hohonet: 360 indoor holistic understanding with latent horizontal
  features.
\newblock In {\em Proceedings of the IEEE/CVF Conference on Computer Vision and
  Pattern Recognition}, pages 2573--2582, 2021.

\bibitem{wang2018self}
Fu-En Wang, Hou-Ning Hu, Hsien-Tzu Cheng, Juan-Ting Lin, Shang-Ta Yang, Meng-Li
  Shih, Hung-Kuo Chu, and Min Sun.
\newblock Self-supervised learning of depth and camera motion from 360 videos.
\newblock In {\em Asian Conference on Computer Vision}, pages 53--68. Springer,
  2018.

\bibitem{wang2020bifuse}
Fu-En Wang, Yu-Hsuan Yeh, Min Sun, Wei-Chen Chiu, and Yi-Hsuan Tsai.
\newblock Bifuse: Monocular 360 depth estimation via bi-projection fusion.
\newblock In {\em Proceedings of the IEEE/CVF Conference on Computer Vision and
  Pattern Recognition}, pages 462--471, 2020.

\bibitem{gmvs2019}
Qingshan Xu and Wenbing Tao.
\newblock Multi-scale geometric consistency guided multi-view stereo.
\newblock In {\em Proceedings of the IEEE/CVF Conference on Computer Vision and
  Pattern Recognition (CVPR)}, June 2019.

\bibitem{yao2018mvsnet}
Yao Yao, Zixin Luo, Shiwei Li, Tian Fang, and Long Quan.
\newblock Mvsnet: Depth inference for unstructured multi-view stereo.
\newblock In {\em Proceedings of the European Conference on Computer Vision
  (ECCV)}, pages 767--783, 2018.

\bibitem{zhang2019orientation}
Chao Zhang, Stephan Liwicki, William Smith, and Roberto Cipolla.
\newblock Orientation-aware semantic segmentation on icosahedron spheres.
\newblock In {\em Proceedings of the IEEE/CVF International Conference on
  Computer Vision}, pages 3533--3541, 2019.

\bibitem{zhao2020towards}
Wang Zhao, Shaohui Liu, Yezhi Shu, and Yong-Jin Liu.
\newblock Towards better generalization: Joint depth-pose learning without
  posenet.
\newblock In {\em Proceedings of the IEEE/CVF Conference on Computer Vision and
  Pattern Recognition}, pages 9151--9161, 2020.

\bibitem{zhou2017unsupervised}
Tinghui Zhou, Matthew Brown, Noah Snavely, and David~G Lowe.
\newblock Unsupervised learning of depth and ego-motion from video.
\newblock In {\em Proceedings of the IEEE conference on computer vision and
  pattern recognition}, pages 1851--1858, 2017.

\bibitem{zioulis2018omnidepth}
Nikolaos Zioulis, Antonis Karakottas, Dimitrios Zarpalas, and Petros Daras.
\newblock Omnidepth: Dense depth estimation for indoors spherical panoramas.
\newblock In {\em Proceedings of the European Conference on Computer Vision
  (ECCV)}, pages 448--465, 2018.

\bibitem{zou2018layoutnet}
Chuhang Zou, Alex Colburn, Qi Shan, and Derek Hoiem.
\newblock Layoutnet: Reconstructing the 3d room layout from a single rgb image.
\newblock In {\em Proceedings of the IEEE Conference on Computer Vision and
  Pattern Recognition}, pages 2051--2059, 2018.

\end{thebibliography}
}

\end{document}